\documentclass[twocolumn]{article}

\usepackage{PRIMEarxiv}

\usepackage[utf8]{inputenc} 
\usepackage[T1]{fontenc}    
\usepackage{hyperref}       
\usepackage{url}            
\usepackage{booktabs}       
\usepackage{amsfonts}       
\usepackage{nicefrac}       
\usepackage{microtype}      
\usepackage{fancyhdr}       
\usepackage{graphicx}       
\graphicspath{{media/}}     

\usepackage{enumitem}
\setitemize{leftmargin=*}
\usepackage{caption}
\usepackage{endnotes}
\usepackage{amsmath}
\usepackage{amssymb}
\usepackage[utf8]{inputenc} 
\usepackage[T1]{fontenc}    
\usepackage{siunitx}
\DeclareSIUnit\pixel{px}
\usepackage{todonotes}
\usepackage{csquotes}
\usepackage{mathtools}
\usepackage{algorithm}
\usepackage{algpseudocode}

\usepackage{float}
\usepackage[super]{nth}
\usepackage{pgf}
\usepackage{pgfplots}
\usepackage{subcaption}
\usepackage{tikz}
\usepackage[format=plain, indention=1.5em, labelfont={bf}, justification=justified]{caption}
\usepackage{setspace}
\usepackage{authblk}

\makeatletter

\let\c@table\c@figure
\makeatother 

\newcommand\hide[1]{}
\newboolean{show_notes}
\setboolean{show_notes}{true}
\newcommand\note[1]{\ifthenelse{\boolean{show_notes}}{\textcolor{red}{\textbf{Note: }#1}}{\hide{#1}}}

\title{Agricultural Plant Cataloging and Establishment of a Data Framework from UAV-based Crop Images by Computer Vision
}

\author[*,1,2]{\textbf{Maurice Günder}}
\author[3]{\textbf{Facundo R. Ispizua Yamati}}
\author[4]{\textbf{Jana Kierdorf}}
\author[4,5]{\textbf{Ribana Roscher}}
\author[3]{\textbf{Anne-Katrin Mahlein}}
\author[1,2]{\textbf{Christian Bauckhage}}

\affil[1]{Fraunhofer Institute for Intelligent Analysis and Information Systems IAIS, Schloss Birlinghoven, 53757 Sankt Augustin, Germany}
\affil[2]{Institute for Computer Science III, University of Bonn, Friedrich-Hirzebruch-Allee 5, 53115 Bonn, Germany}
\affil[3]{Institute for Sugar Beet Research (IfZ), Holtenser Landstraße 77, 37079, Germany}
\affil[4]{Institute for Geodesy and Geoinformation, University of Bonn, Niebuhrstraße 1a, 53113 Bonn, Germany}
\affil[5]{Department of Aerospace and Geodesy, Data Science in Earth Observation, Technical University of Munich, Lise-Meitner-Straße 9, 85521 Ottobrunn, Germany}

\affil[*]{Corresponding author, e-Mail: \url{mguender@uni-bonn.de}}

\begin{document}

\twocolumn[
  \begin{@twocolumnfalse}
    \maketitle
    \begin{abstract}
        UAV-based image retrieval in modern agriculture enables gathering large amounts of spatially referenced crop image data. In large-scale experiments, however, UAV images suffer from containing a multitudinous amount of crops in a complex canopy architecture. Especially for the observation of temporal effects, this complicates the recognition of individual plants over several images and the extraction of relevant information tremendously.
        In this work, we present a hands-on workflow for the automatized temporal and spatial identification and individualization of crop images from UAVs abbreviated as \enquote{cataloging} based on comprehensible computer vision methods. We evaluate the workflow on two real-world datasets. One dataset is recorded for observation of Cercospora leaf spot\textemdash a fungal disease\textemdash in sugar beet over an entire growing cycle. The other one deals with harvest prediction of cauliflower plants. The plant catalog is utilized for the extraction of single plant images seen over multiple time points. This gathers large-scale spatio-temporal image dataset that in turn can be applied to train further machine learning models including various data layers. 
        The presented approach improves analysis and interpretation of UAV data in agriculture significantly. By validation with some reference data, our method shows an accuracy that is similar to more complex deep learning-based recognition techniques. Our workflow is able to automatize plant cataloging and training image extraction, especially for large datasets.
    \end{abstract}
    \end{@twocolumnfalse}
    ]

\clearpage

\section{Background}\label{chap:background}

\begin{figure*}[t]
    \centering
    \includegraphics[width=\textwidth]{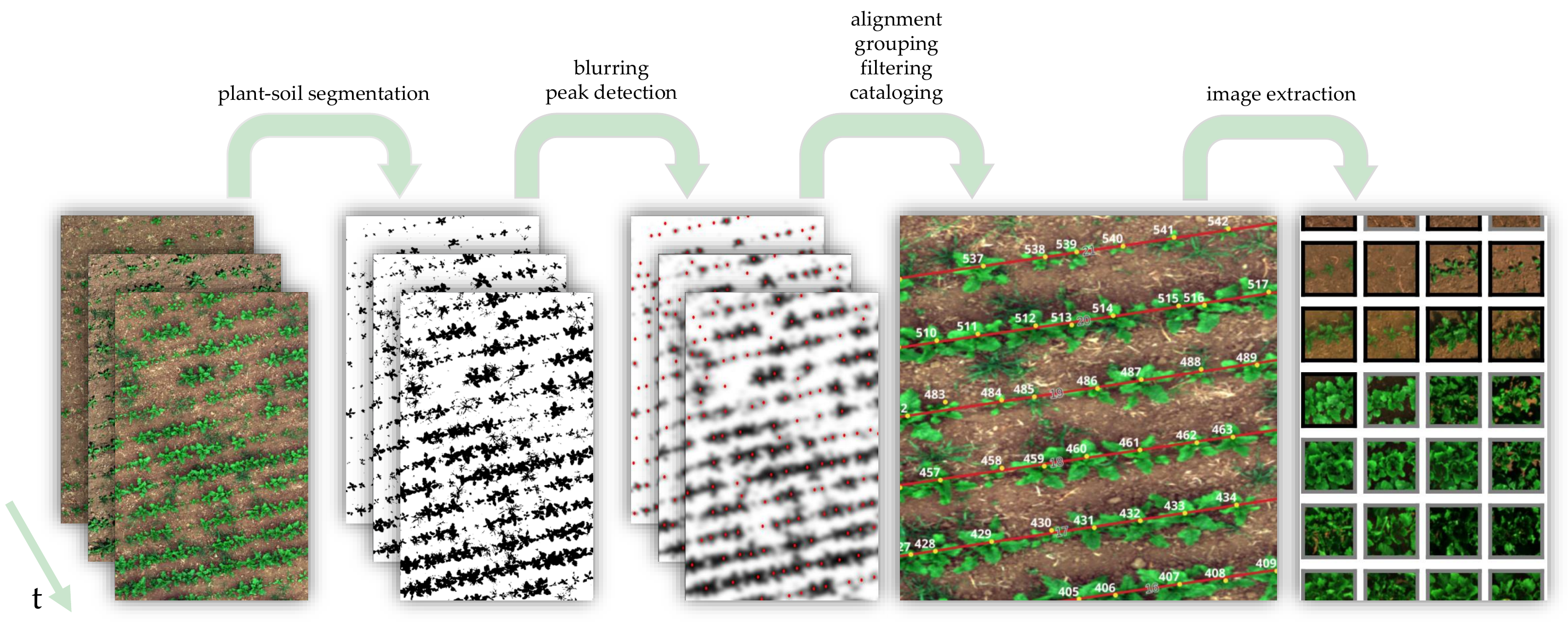}
    \caption{\textbf{Overview of our proposed plant cataloging and image extraction workflow.} The multi-channel UAV images with different acquisition dates are processed for plant-soil segmentation. Afterwards, adaptive Gaussian blur filtering helps to locate the plant positions more precisely via peak finding. Subsequently, the peaks are grouped by time and further manipulation like seeding line detection leads to the plant catalog. Once this catalog is available, one can extract image time series data of single plants.}
    \label{fig:overview}
\end{figure*}
The use of Unmanned Aerial Vehicles (UAVs) is one of the main drivers of modern precision agriculture. Equipped with cameras and other sensors like LiDAR, UAVs can be used for diverse non-invasive in-field analyses and observation tasks~\cite{intro_remote_sensing}. Analyzing and interpreting resulting data with Machine Learning and Pattern Recognition methods has the potential to gain economical and ecological efficiency, which is why computational intelligence is increasingly applied in agricultural contexts~\cite{pr_agri_bauckhage, dl_in_agri}. Unlike satellite-based remote sensing~\cite{intro_rs_overview}, UAV-based remote sensing offers low-cost solutions for private usage and individual applications. There are many conceivable use cases like plant species segmentation~\cite{intro_rs_plant_species_segmentation}, multi-sensor plant analyses~\cite{intro_rs_multisensor1, intro_rs_multisensor2, intro_cauliflower_phenotyping}, or automated plant counting~\cite{intro_rs_automated_plant_counting}. Among those, several tasks are based on investigations on individual plants. This is challenging for applications in real-world farming where the fields are densely seeded, in contrast to research cases where plants possibly could have larger distances. 
We will make use of the cost-effective UAV imaging which enables having multiple images during a complete growing season and combine this information for localizing plants in a spatio-temporal way. Reasons and use cases for cataloging and detecting plants in the field are manifold. On the one hand, it allows determining the number of plants in the field accurately. On the other hand, it enables the extraction of traits of great interest, such as the distance between plants and plant density. Individualization and identification of each plant in the field and the opportunity of retrieving individual plants again in a time series allow generating a vast database of annotations. The availability of this data framework enables to train more robust machine learning models for disease severity analysis, to mention just one possible use case. 

To observe individual plants, the exact positions of the plants in the image are required. For the moment, it is possible in two arduous and time-consuming ways: (1) by georeferencing the plants with sub-centimeter accuracy directly on the field by using a GPS receiver with real-time kinematic positioning (RTK) or (2) after image acquisition and orthorectification by manual annotating with a geographic information system (GIS). When taking time series images, the perfect alignment of these images requires the use of well-known and measured points, called ground control points (GCPs)~\cite{intro_gcp}. If the GCPs are correctly georeferenced, they allow allocation of the images not only in a local but also in a global coordinate system, such as WGS84. This enables pixel-precise transferability between images from different locations and time points. The localization of a plant is thus only necessary at one point in time. The determined absolute coordinate remains identical at every point in time, since the plant does not move in space. Thus, a simple extraction of the plant at different points in time is possible. Overlapping of neighboring plants does not affect localization in this approach if the position is calculated in an early growth stage. 
If no georeferencing of the data is available, e.g. because no access to measuring devices or GCPs is possible, another possibility is to match the images using plant position detection based on single images. In a further step, those detection points can be registered to each other. However, dealing with overlaps of plants in later developmental stages is challenging in this approach.

Once the different time points are aligned, it is feasible to extract time series for each individual plant. Time series allow the analysis of the plant growth and especially the assessment of the plant traits and their development over time. Based on time series, it is possible to identify the individual development of each plant over time. This spatio-temporal information enables optimization of crop management. Another aspect is the detection of stress factors and disease development, and how it influences the growth of the affected plant over time. 

In this work, we introduce a complete workflow for plant identification and individualization\textemdash further also referred to as \enquote{cataloging}\textemdash of single plants. Figure~\ref{fig:overview} summarizes our proposed workflow with its main steps. We will evaluate the workflow on ground truth data and give possible use case suggestions.

\section{Data Description}\label{chap:data}

In order to show, that our workflow can be used on a variety of UAV image data, we evaluate our approach on two agricultural datasets with different contexts of use. However, the transfer to other datasets and use cases is possible. On the one hand, a dataset from sugar beet for research in the context of Cercospora leaf spot disease spread is used. As this disease is critically affecting the yield of sugar beet farming, several phenotyping studies with UAVs and Unmanned Ground Vehicles (UGVs) have already been done~\cite{intro_cercospora_classification, intro_cercospora_phenotyping}. On the other hand, a dataset is used that monitors the development and harvest time of cauliflower. Such data can be used to extract phenotypic traits such as diameter or height of cauliflower plants or their head~\cite{intro_cauliflower_phenotyping, intro_temporal_prediction}. Additionally, plant disease research in general is an excellent use case for this workflow since it requires having individual plant information throughout many growth stages~\cite{intro_plant_disease1, intro_plant_disease2, intro_plant_disease3}. Thus, a spatio-temporal individualization of the investigated plants is crucial. Especially for large-scale investigations, an approach as automatized as possible is desirable.

Some challenges for both data sets are for example the varying exposure due to the different data acquisition times while the whole growing season and associated different weather conditions. Occurring weeds can have positive as well as negative effects. On the one hand, it simplifies tasks such as registration, since weeds are stationary and thus serve as markers. On the other hand, the growth of weeds affects tasks such as plant detection, since a distinction must be made beforehand between weeds and crops. Another challenge occurs when the plants are so large that the canopy of the plants is closed. This complicates the separation of different plant instances and the search for distinctive points in the field, which is often made possible by the contrast between soil and plants.

\subsection{Sugar beet dataset}

The sugar beet data set was used to develop and evaluate our workflow steps which consists of multispectral drone (UAV) images over time.

\begin{figure}[t]
    \centering
    \includegraphics[width=0.9\columnwidth]{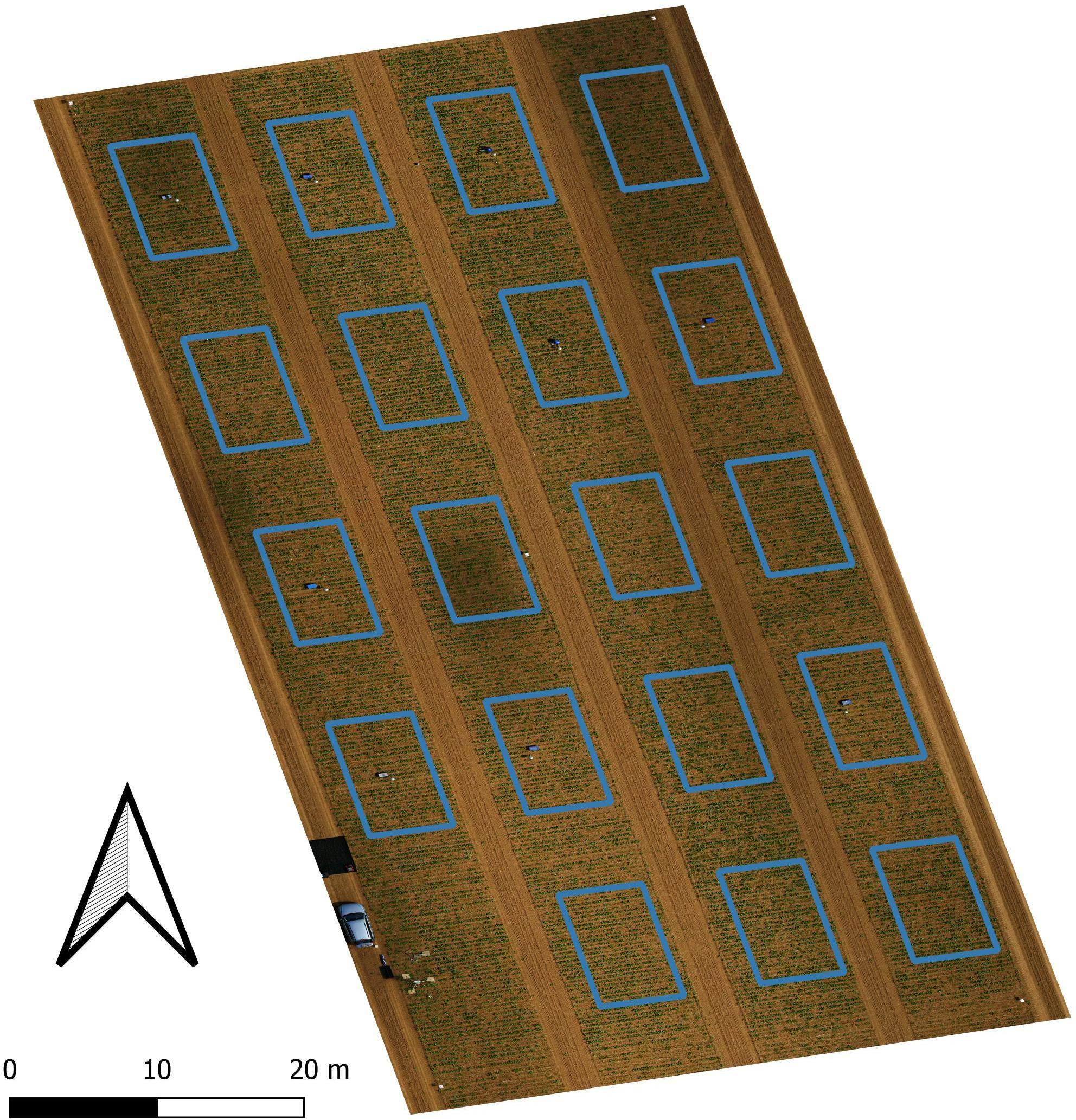}
    \caption{\textbf{Overview of Cercospora leaf spot experimental field 2020 in Göttingen, Germany.} The 19 marked areas correspond to the plots under analysis and undergo different treatments like disease inoculation and fungicide application.The measurements were conducted throughout the full growing season.}
    \label{fig:cercospora_fields}
\end{figure}

Figure~\ref{fig:cercospora_fields} shows the location of the plots in the trial field. It was conducted in 2020 near Göttingen, Germany (\ang{51;33;}N \ang{9;53;}E) on a weekly basis at \num{24} dates during the complete growing season from May \nth{7} to October \nth{12}, 2020. On these, one variety of sugar beet plants that is susceptible against CLS was selected\textemdash the \textit{Aluco}\footnote{SESVanderHave} variety. The sugar beet was sown on April \nth{6}, 2020, with an inter-row distance of \SI{48}{\centi\meter}, an intra-row distance of \SI{18}{\centi\meter}, and an expected seeding rate of approximately \num{110000} plants per hectare.

For the aerial data collection, the quadrocopter \textit{DJI Matrice 210}\footnote{SZ DJI Technology Co., Ltd., Shenzhen, China\label{fn:quadrocopter}} was used. The position of the UAV during the flight was determined by GPS and GLONASS and corrected by the \textit{D-RTK 2 Mobile Station}\textsuperscript{\ref{fn:quadrocopter}}.
The camera mounted is a \textit{Micasense Altum} multispectral camera\footnote{MicaSense, Inc., USA} with 6 bands: blue (\SI{475}{\nano\meter} center, \SI{32}{\nano\meter} bandwidth), green (\SI{560}{\nano\meter} center, \SI{27}{\nano\meter} bandwidth), red (\SI{668}{\nano\meter} center, \SI{14}{\nano\meter} bandwidth), red edge (\SI{717}{\nano\meter} center, \SI{12}{\nano\meter} bandwidth), near infrared (\SI{842}{\nano\meter} center, \SI{57}{\nano\meter} bandwidth), and long-wave thermal infrared (\SIrange{8000}{14000}{\nano\meter}) with a \textit{Downwelling Light Sensor} (DLS2). The ground resolution was set to \SI{4}{\milli\meter\per\pixel}. Side-overlap and forward-overlap were set to \SI{75}{\percent}. The drone flew in the range around \SI{14}{\meter} above ground level with a speed of \SI{0.5}{\meter\per\second} and a resolution of $\SI{4}{\milli\meter}\times\SI{4}{\milli\meter}$ per pixel. 

In addition, ground control points (GCPs) were set up at the corners and in the middle of the test arrangement for the multitemporal monitoring in the sense of georeferencing and correcting the collected data. \textit{Agisoft Metashape Professional}\footnote{Version 1.6.3.10732} software has been used to create orthomosaic images. QGIS\footnote{Version 3.16}~\cite{qgis} was used to delimit and select the area to be analyzed, resulting in 19 plots with an equal area of \SI{172}{\meter\squared} each. 

The experiment present presented three levels: (1) control with fungicide, (2) inoculated with fungicide, and (3) inoculated without fungicide. Plots of inoculated treatments were infected manually with CLS-diseased sugar beet air-dried leaf material. 

The first image set was generated in an experimental field focused on detecting of Cercospora leaf spot disease (CLS). CLS is caused by \textit{Cercospora beticola} Sacc. and is one of the most important leaf diseases of sugar beet worldwide. The actual aim of the experiment is to be able to develop models for an accurate and early detection by monitoring of the disease with information from optical and environmental sensors. However, common models are affected by the interaction between plants and the reaction of each plant to the environment. To achieve more accurate and georeferenced information about the plant and its interaction with the pathogen, the need arises to determine the individual position of each plant throughout the development of the growing season.

For the quality evaluation of our workflow, some fields are annotated by human experts, preferably in the earlier stages where the plants are locally distinguishable, by marking the central position of each visible plant.

\subsection{Cauliflower Dataset}\label{chap:cauliflowerDataset}
The cauliflower dataset serves for validation of our workflow. It consists of a time series of RGB UAV images of a \SI{0.6}{\hectare} sized cauliflower field of the Korlanu variety taken on a weekly basis between July, \nth{28} and November \nth{2}, 2020 near Bonn, Germany (\ang{50;46;}N \ang{6;57;}E). The plants are planted with an intra-row distance of \SI{50}{\centi\meter} and inter-row distance of \SI{60}{\centi\meter}. The cauliflower field was monitored once a week throughout the growing season. As with the sugar beet dataset, the images of the cauliflower dataset were processed into orthophotos using \textit{Agisoft Metashape Professional} software. The flight altitude of the \textit{DJI Matrice 600}\textsuperscript{\ref{fn:quadrocopter}} hexacopter was \SI{10}{\meter} and with a \textit{Sony A7 rIII} RGB camera, we obtain a ground resolution of the orthophotos of roughly $\num{1.5}\times \SI{1.5}{\milli\meter\per\pixel}$. 21 GCPs were distributed throughout the field and are measured using real-time kinematic positioning (RTK).

The idea behind the acquisition of the dataset is to observe cauliflower plants over its development and to develop models based on the acquired time series that predict, for example, the harvest time of each plant in the field or reflect the maturity state of the plant. Harvesting cauliflower is very laborious, so over several days field workers walk across the field and manually harvest ripe cauliflower heads. In this process, the size of the cauliflower heads is the most crucial plant trait that determines the harvest. Thanks to georeferenced monitoring, it is attempted to help the farmer to optimize his fertilization and pest control as well as to work more economically, in that the field workers already know in advance which plants are to be harvested and do not have to check them individually.
  
\section{Plant Extraction Workflow}

In this section, we will describe the full plant extraction workflow in detail. For each workflow step, we present plots on different excerpts from the sugar beet dataset in order to visualize the results.

\subsection{Finding Individual Plants}\label{chap:plant_positions}

As the first step of plant cataloging, the plant positions are determined for each acquired image, i.e. for each plot at each acquisition date. For this, we perform a segmentation between soil and plants. Common approaches use \textit{Vegetation Indices} (VIs) as a preprocessing for the image information~\cite{intro_rs_plantsegment_survey}. There are pure RGB-based VIs as well as VIs including multi- or hyperspectral information. Due to the multispectral acquisitions, we have available in the sugar beet dataset, the contrast between plants and soil could be increased compared to pure RGB images. However, our experiments showed that for our dataset, including multispectral information, did not result in better results than with pure RGB information. This is an evidence that our workflow performs well even for use cases where beyond-RGB imaging is not feasible.

\subsubsection{Vegetation Index}\label{sec:veg_ind}

\begin{figure}[t]
    \centering
    \includegraphics[width=\columnwidth]{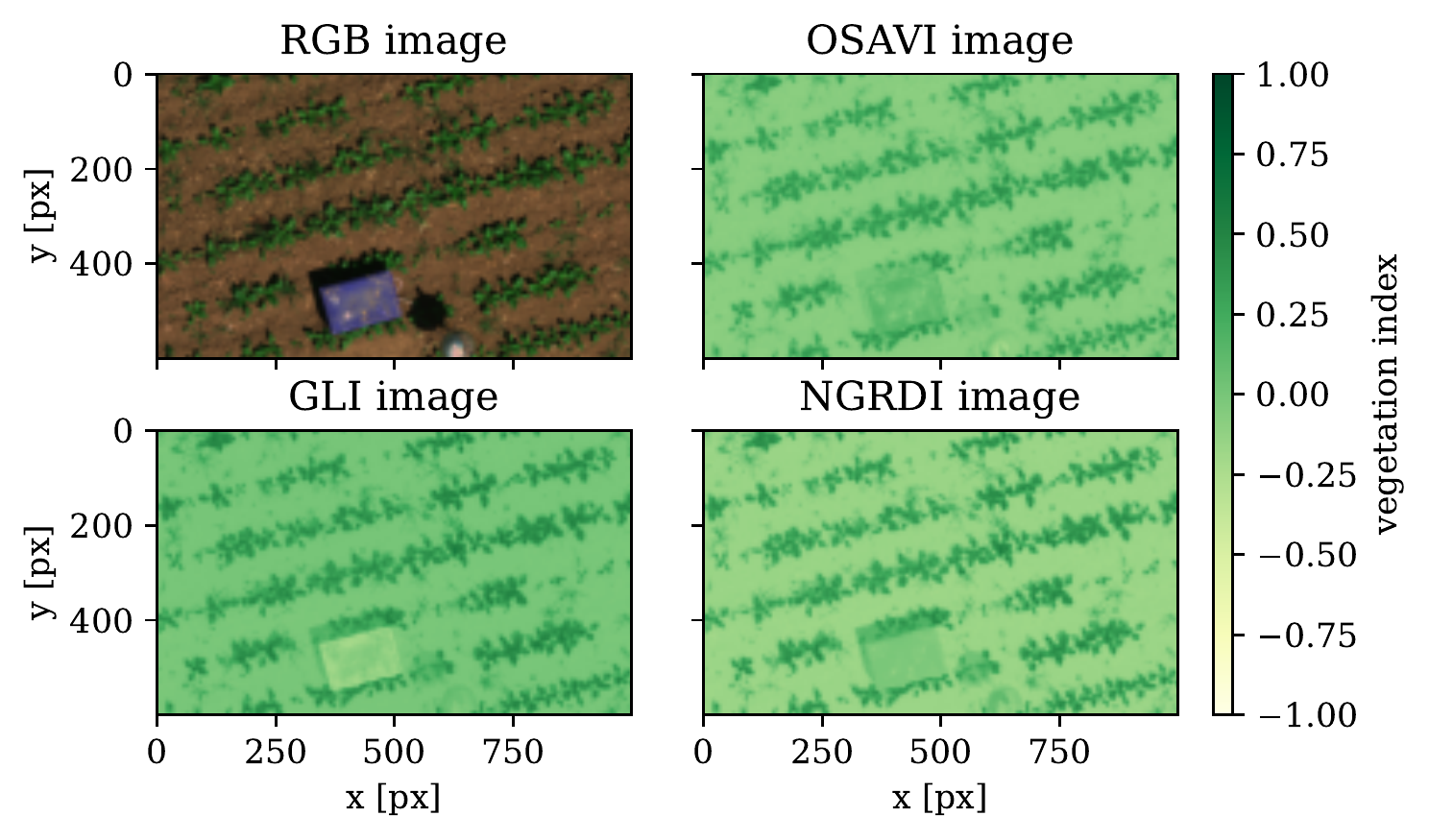}
    \caption{\textbf{Different VIs for discrimination between plants and soil.} The RGB image information (upper left) and three Vegetation Indices are shown. The OSAVI as a multispectral index (upper right) as well as the RGB-only indices GLI (lower left) and NGRDI (lower right) can be calculated for the multispectral sugar beet dataset. Using multispectral information, OSAVI is less susceptible to foreign objects in the field than RGB-only indices.}
    \label{fig:vi_example}
\end{figure}

Using spectral information to condense the information in a one-channel image is the main concept of VIs. Besides reducing the image dimension, the goal behind introducing the VI is to enhance the contrast between soil and plants compared with a standard RGB image, for instance. There are plenty of publications of different vegetation indices so far~\cite{veg_ind_survey1,veg_ind_survey2}. However, they all combine different spectral channels by a certain calculation specification. In addition, some indices use (empirically motivated) parameters to further optimize them for different use cases.
Since we want to demonstrate that our plant extraction method works for pure RGB information already, we stick to RGB-based VIs. Two indices that turn out to work well in our experiments are the \textit{\textbf{G}reen \textbf{L}eaf \textbf{I}ndex}~\cite{gli}
\begin{equation*}
    GLI_i = \frac{2G_i-R_i-B_i}{2G_i+R_i+B_i}\,,
\end{equation*}
as well as the \textit{\textbf{N}ormalized \textbf{G}reen/\textbf{R}ed \textbf{D}ifference \textbf{I}ndex}~\cite{ngrdi}
\begin{equation*}
    NGRDI_i = \frac{G_i-R_i}{G_i+R_i}\,.
\end{equation*}
These indices map the information of red ($R$), green ($G$), and blue ($B$) channels of each image pixel $i$ onto a single value in the interval $[-1,1]$. If there is multispectral data available, nevertheless, one may use the \textit{\textbf{O}ptimized \textbf{S}oil \textbf{A}djusted \textbf{V}egetation \textbf{I}ndex}~\cite{osavi} defined as
\begin{equation*}
    OSAVI_i = \frac{NIR_i-R_i}{NIR_i+R_i+Y}\,,
\end{equation*}
where $Y>0$ is an empirical parameter. In this work, we use $Y=\num{0.6}$. Although it is not necessary to work with multispectral VIs in this stage, they can further discriminate plants from foreign objects in the UAV image data. Figure~\ref{fig:vi_example} shows the mentioned VIs applied on an example image snippet taken from the sugar beet dataset. The remarkable points are that all three VIs can handle shadows fairly well, but the OSAVI is slightly better in ignoring foreign objects.

Note that, surely, also other Vegetation Indices that produce substantial contrast between plants and soil are conceivable at this point. However, we restrict us to the RGB-only VI images for the further plant finding procedure at this place. They are named as $\mathbf{V}(t)$ for each acquisition date $t\in \mathcal{T}$ in the set of given acquisition dates $\mathcal{T}$. The single pixel values are named by $v_i$.

\subsubsection{Plant-soil Segmentation}\label{sec:plant_seg}

In the next step, the soil has to be discriminated from the plant area. Thus, we will perform a segmentation.

As section \textit{\nameref{sec:veg_ind}} may already have suggested, the ratio between plant and soil coverage in the images is highly variable for an image series during a complete growing season. Therefore, it is important to keep the growing state\textemdash or the plant size\textemdash in mind for each image. Thus, a convenient measure is the cover ratio $c$ defined by
\begin{equation*}
    c = \frac{1}{N_\text{px}} \sum\limits_i^{N_\text{px}} \chi_{\{v_i \geq \vartheta\}}\,
\end{equation*}
where $v_i$ is the VI value for the $i$-th of $N_\text{px}$ total pixels and $\vartheta$ is a predefined threshold value. $\chi_{\{\cdot\}}$ is the indicator function, which is $1$ if the condition given in the index is met and $0$ else.
Descriptively, the cover ratio is just the percentage of pixels above threshold.
For each image, this cover ratio $c$ is calculated. For our two preferred VIs, we set the thresholds to $\vartheta_\text{GLI}=\num{0.2}$ and $\vartheta_\text{NGRDI}=\num{0}$, respectively. For OSAVI, the threshold would be set to $\vartheta_\text{OSAVI}=\num{0.25}$.

Since the cover ratio is a proxy for the plant size, it enables us to roughly classify the different growing stages, especially the edge cases of very small ($c\to\num{0}$) or large ($c\to\num{1}$) plants. 

\begin{figure}[t]
    \centering
    \begin{subfigure}{0.9\columnwidth}
        \includegraphics[width=\columnwidth]{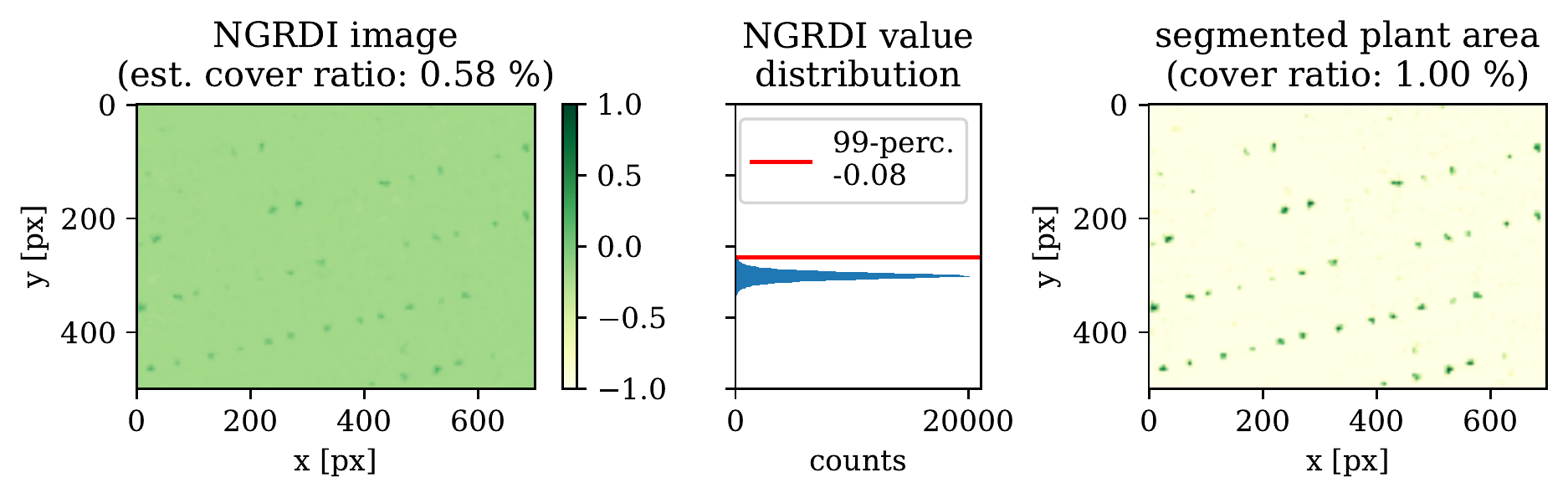}
        \subcaption{}\label{fig:plant_seg_1}
    \end{subfigure}
    \begin{subfigure}{0.9\columnwidth}
        \includegraphics[width=\columnwidth]{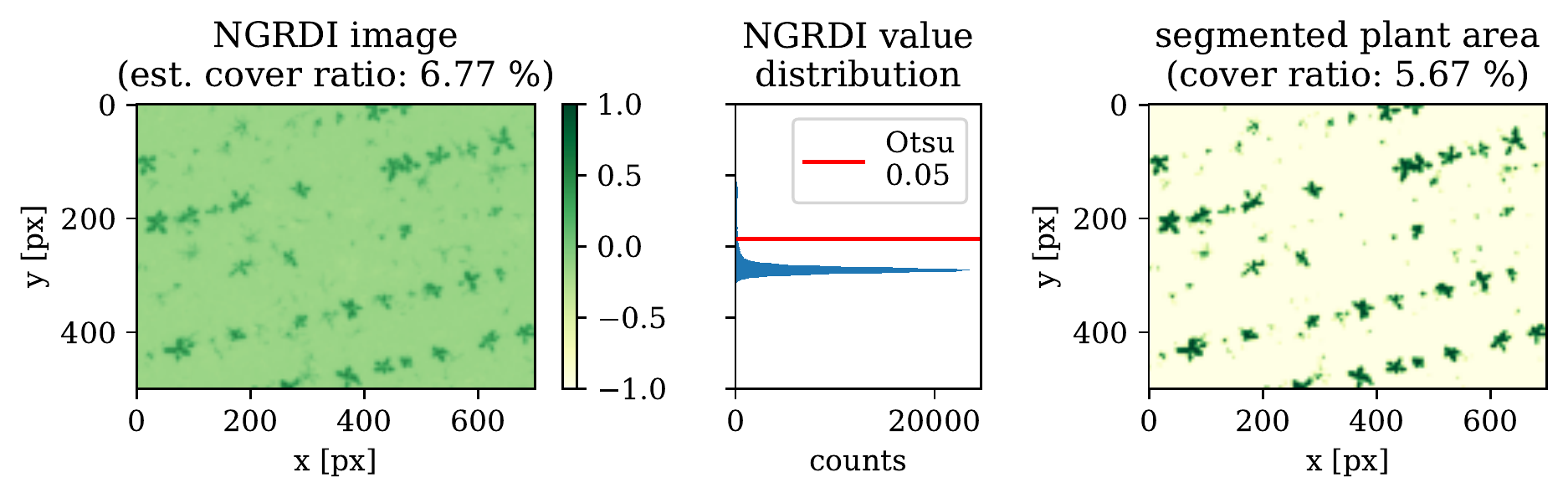}
        \subcaption{}\label{fig:plant_seg_2}
    \end{subfigure}
    \begin{subfigure}{0.9\columnwidth}
        \includegraphics[width=\columnwidth]{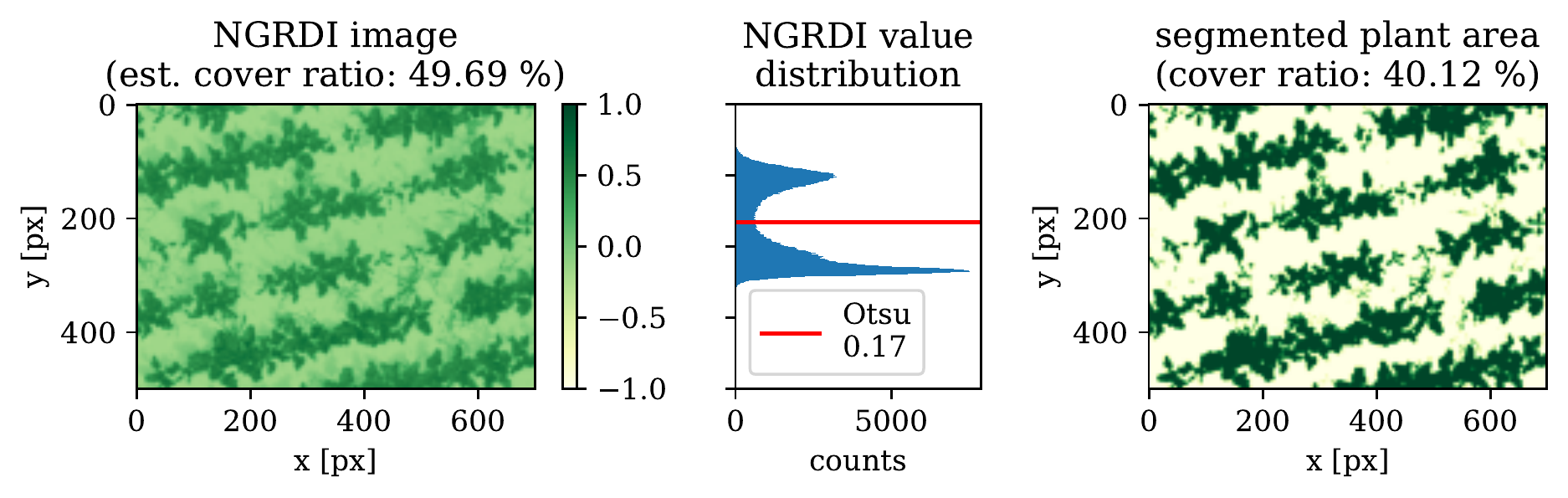}
        \subcaption{}\label{fig:plant_seg_3}
    \end{subfigure}
    \begin{subfigure}{0.9\columnwidth}
        \includegraphics[width=\columnwidth]{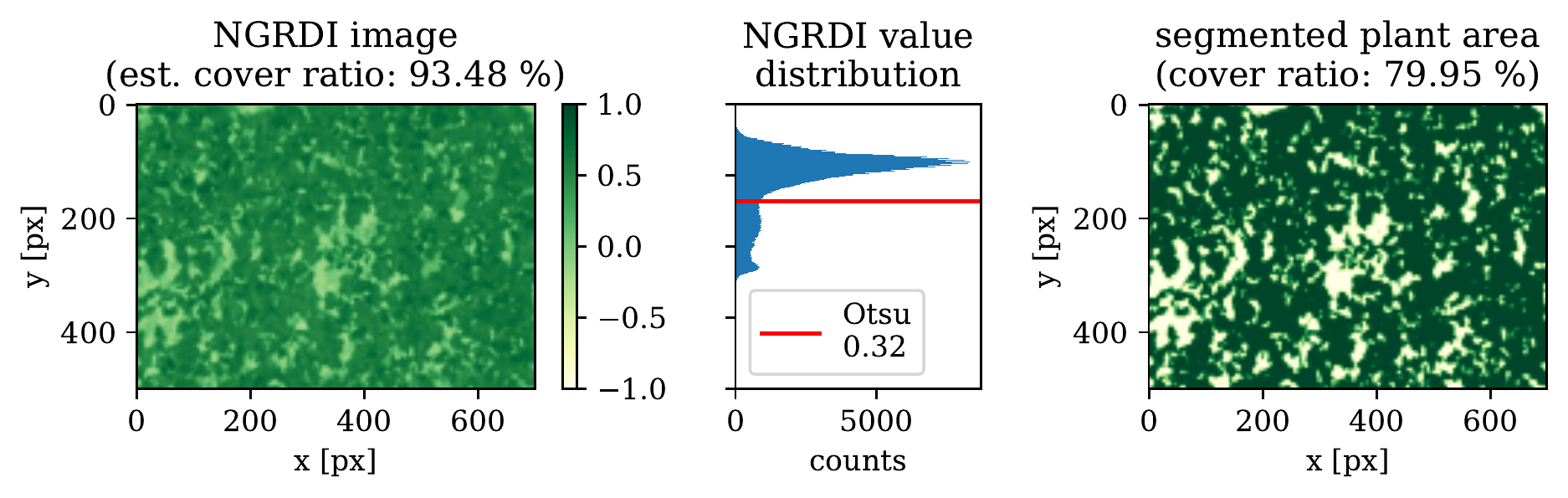}
        \subcaption{}\label{fig:plant_seg_5}
    \end{subfigure}
    \caption{\textbf{Plant-soil segmentation for different growing stages.} The left plots show the NGRDI image. The middle histograms show the distributions of pixel values and the applied threshold for the segmentation\textemdash as seen in the right plots. In (\subref{fig:plant_seg_1}), the estimated cover ratio is below \SI{1}{\percent} so that the 99-percentile value is used as threshold. In cases depicted in (\subref{fig:plant_seg_2}) and (\subref{fig:plant_seg_3}), Otsu's method is applied. Until roughly \SI{75}{\percent} cover ratio, single plants and seeding lines are visible. Images beyond that limit like (\subref{fig:plant_seg_5}) are not used for further analysis.}
    \label{fig:plant_seg}
\end{figure}

As a more sophisticated threshold technique, we use Otsu's method~\cite{otsu_thresholding}. The main idea is to find the threshold that minimizes the intra-class variance of two classes to be considered as fore- and background. For intermediately covered fields, this method leads to a reliable discrimination between soil and plants. However, it has weaknesses for the following extreme cases. For very sparsely covered fields ($c<\SI{1}{\percent}$), Otsu's method usually results in a threshold that is approximately at the most frequent value\textemdash i.e.~the mode\textemdash of the Vegetation Index values. This is because the small amount of \enquote{plant pixels} simply disappears in the value distribution, which then has a single peak structure. In those images, the most frequent value is at soil level; thus, we would obtain much noise by using Otsu's threshold and would overestimate the cover ratio calculated by the resulting segmentation mask. We bypass this problem by setting the threshold to the 99-percentile of the VI distribution, which obviously results in a modified cover ratio of exactly $\SI{1}{\percent}$. For cover ratios above circa \SI{75}{\percent}, single plants or seeding lines are not properly distinguishable anymore. We therefore do not use those stages for segmentation. Figure~\ref{fig:plant_seg} summarizes the findings by some exemplary field sectors during different growing stages.

\subsubsection{Excursus: Growth Function}\label{sec:growth_function}

\begin{figure}[t]
    \centering
    \includegraphics[width=\columnwidth]{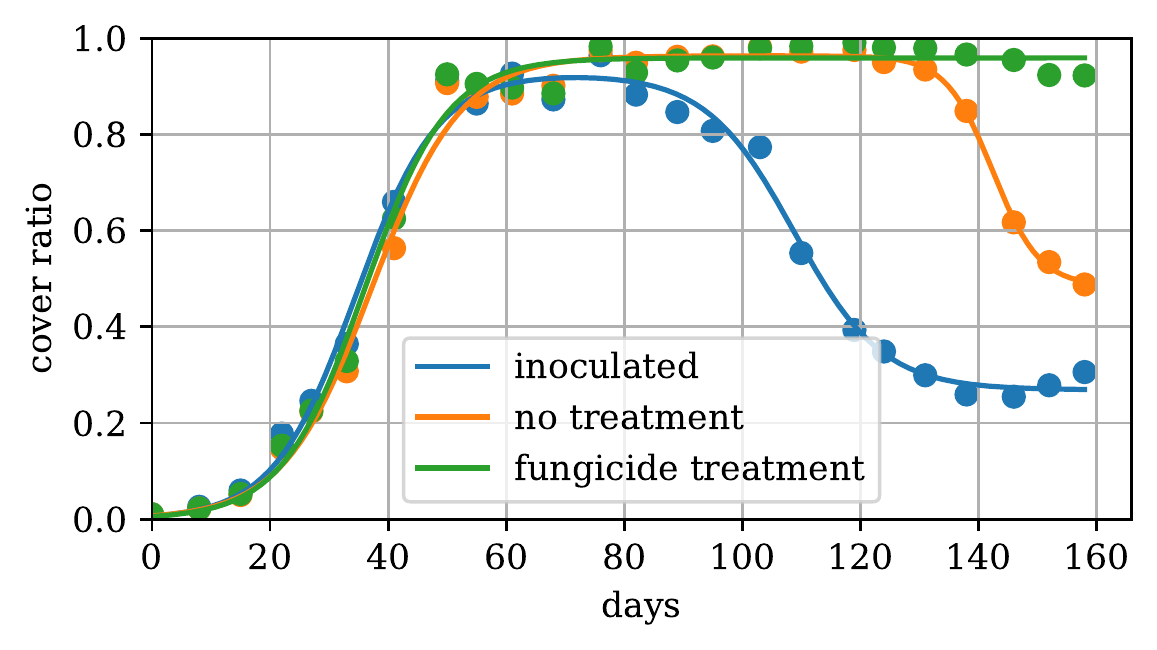}
    \caption{\textbf{Growth function curve fits.} The cover ratio estimates vs. days since first acquisition is fitted with the growth function~\eqref{eq:grow_func}. The shape of the growth function is different for the 3 categories. The blue curve shows an inoculated field. After the growing phase, the dying process due to the disease starts and saturates in the latest acquisitions. In case of the untreated field (yellow curve), the natural dying process is not saturated. In the fungicide treated field shown by the green curve, no dying phase is visible. Thus, the second dying term of the fit function is not present, in contrast to the first two cases.}
    \label{fig:grow_func}
\end{figure}

If enough images during a complete growing season are acquired, we can estimate a functional relation between the single estimated cover ratio. Empirically and plotted against the acquisition date, the cover ratio follows a saturated exponential function until the plants are dying or suffering diseases. From then on, the cover ratio decreases exponentially. This phenomenological considerations motivate the approach for a \textit{growth function} $f(t)$ defined as:
\begin{equation}\label{eq:grow_func}
    f(t) = \frac{g}{1+\exp{(-\lambda_g(t-t_g)})} - \chi_{\{d>0\}}\frac{d}{1+\exp{(-\lambda_d(t-t_d)})}\,.
\end{equation}
Basically, we approximate the growing and dying phase as two independent sigmoid functions. The difference of them is the complete growing function. The growing (dying) slope constants $\lambda_g$ and $g$ ($\lambda_d$ and $d$) as well as the corresponding time offsets $t_g$ ($t_d$) are treated as optimization parameters. By separating cases for positive and non-positive $d$, one can give the optimization process the option to ignore the dying phase if it is not visible. 
For the time $t$, the days since the first acquisition is used. Figure~\ref{fig:grow_func} shows the growth function fit for 3 fields with different treatments.

\subsubsection{Filtering and Peak Finding}\label{sec:filtering_peakfinding}

As segmentation masks are calculated, one can find single plants by applying filtering and peak finding techniques. Recap that we have the segmentation masks (cf.~\textit{\nameref{sec:plant_seg}}) represented as binary images
\begin{equation*}
    \mathbf{B}(t) = \Big\{\chi_{\{v_i \geq \vartheta(t)\}} \Big\}_{i=0}^{N_\text{px}}
\end{equation*}
by applying Otsu's or 99-percentile threshold $\vartheta(t)$ for each VI image pixel $v_i$ and acquisition date $t$. The resulting binary images $\mathbf{B}(t)$ can then be blurred with a Gaussian filter. Mathematically, the images $\mathbf{B}(t)$ are convolved with a 2D Gaussian kernel
\begin{equation*}
    \mathcal{G}(x,y,t) = \frac{1}{2\pi\sigma^2(t)} \exp\left(-\frac{x^2 + y^2}{2\sigma^2(t)}\right)
\end{equation*}
where $\sigma(t)$ is the bandwidth parameter. $x$ and $y$ represent the 2-dimensional image pixel dimensions. Note the time dependency here, since it makes sense to adapt the bandwidth with the size of the plants\textemdash i.e.~given by our cover ratio estimate\textemdash and by introducing an interval of reasonable bandwidth for the given images $[\sigma_\text{min},\sigma_\text{max}]$ in units of image pixels.

As a proxy, the given bandwidth boundaries should roughly represent the size of the plants in the images of the beginning growing stage and maximum growing stage, respectively. Finally, one gets the blurred images $\mathbf{\Tilde{B}}(t)$ by convolution ($\ast$) of the binary images $\mathbf{B}(t)$ with the Gaussian kernel matrix $\mathbf{G}(t)$, hence
\begin{equation*}
    \mathbf{\Tilde{B}}(t) = \mathbf{B}(t)\ast \mathbf{G}(t)\,.
\end{equation*}

Subsequently, a simple peak finder is applied to extract the plant centers. The blurring has the effect, that the plants are visible as blurred quasi-circular objects. In this way, the peak finding algorithm detects the plant center rather than individual leaves. To further improve the results of this peak finding, we can set a minimal peak distance and/or intensity to avoid double detection of bigger plants or still visible weed.

Nevertheless, the outcome of this method is highly dependent on choosing the ranges for binary thresholds and Gaussian filter bandwidth correctly. Therefore, the bandwidths boundaries have to be deliberately adapted to the pixel resolution and the minimal (maximal) plant sizes in the given images.

\begin{figure}[t]
    \centering
    \begin{subfigure}{\columnwidth}
        \includegraphics[width=\columnwidth]{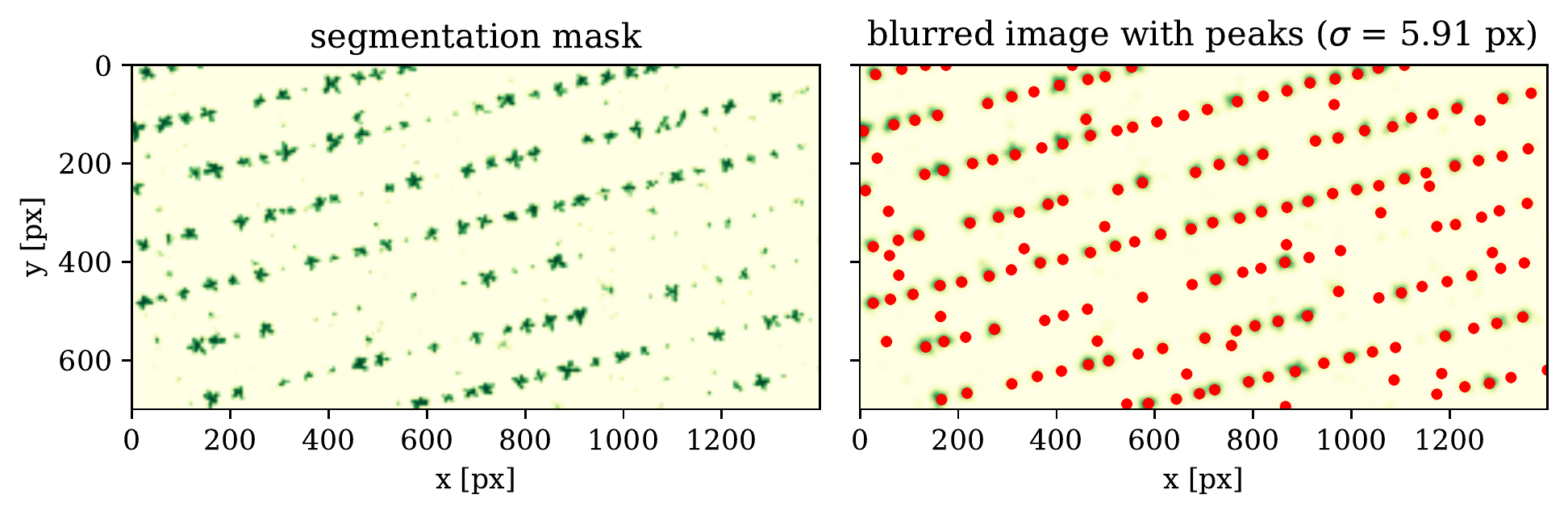}
        \subcaption{}\label{fig:peak_det_1}
    \end{subfigure}
    \begin{subfigure}{\columnwidth}
        \includegraphics[width=\columnwidth]{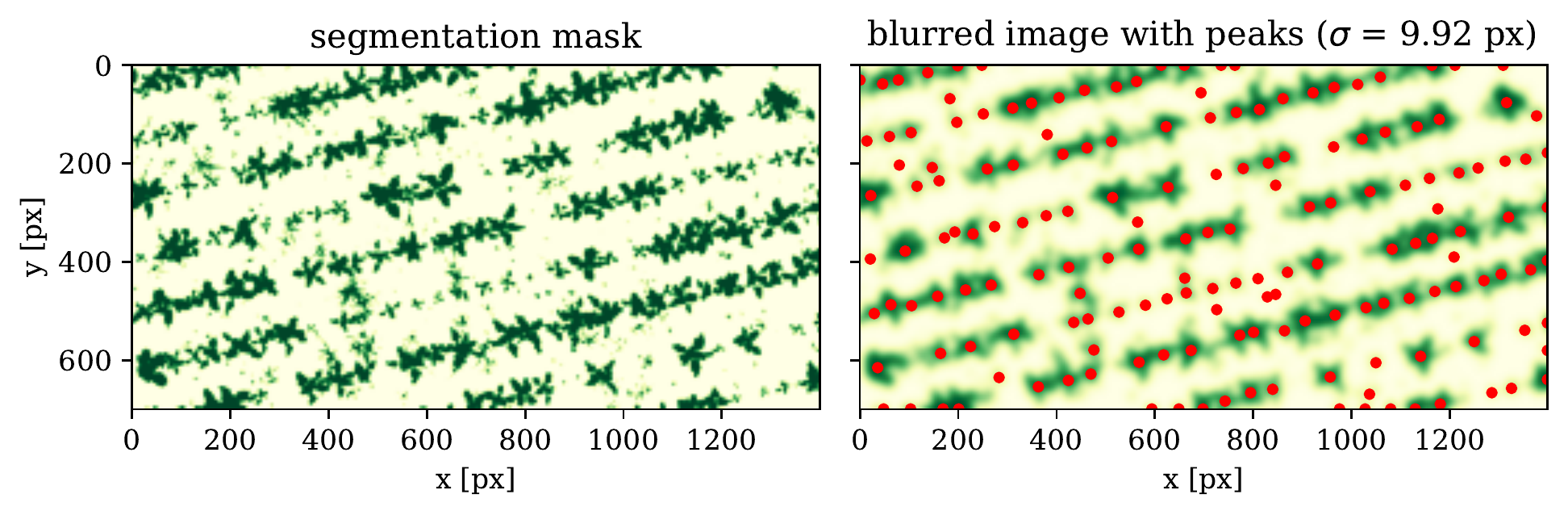}
        \subcaption{}\label{fig:peak_det_2}
    \end{subfigure}
    \caption{\textbf{Plant peaks for different growing stages.} The left plots show the determined segmentation masks for different growing stages of different fields. On the right, the results of the Gaussian blurring with the adaptive bandwidth and the peak detection (red dots) is shown.}
    \label{fig:peak_det}
\end{figure}

Moreover, this method is generally performing significantly better on images with lower plant cover ratios. For higher cover ratios, the plants are no longer spatially divided, which results in misdeterminations or multi-detections. As stated in Figure~\ref{fig:plant_seg}, single plants are hard to discriminate above circa $\SI{75}{\percent}$ cover ratio. Thus, we only perform this method on the lower cover ratio images. Figure~\ref{fig:peak_det} shows plant detection results for different exemplary growing stages. It happens that also unwanted weed is recognized by this method or not every single plant is detected. However, the steps in the following sections minimize those weaknesses by some further filtering and reconstruction methods.

\subsection{Grouping the Plants}\label{chap:plant_grouping}

The steps described in chapter \textit{\nameref{chap:plant_positions}} yield the peak positions for each valid\textemdash i.e.~with sufficiently low cover ratio\textemdash UAV image as pixel coordinates. By using the available geospatial information, the rasterized pixels can be translated into absolute GPS coordinates. Further, we apply a Universal Transverse Mercator (UTM) coordinate system to translate longitude and latitude into a metric scale, e.g. centimeters. This yields metric peak positions
\begin{equation*}
    \mathcal{P}(t) = \Big\{\vec{x}_i(t)\Big\}_{i=0}^{N_p(t)}\,,
\end{equation*}
where $\vec{x}_i(t)$ is the position vector of the $i$-th peak and $N_p(t)$ is the number of detected peaks in the image with acquisition date $t$. Further, we define $\mathcal{T}$ as the set of all acquisition dates and $\mathcal{T}^\ast\subseteq\mathcal{T}$ as the set of all acquisition dates where the peak extraction method was performed. The goal of this section is to group this\textemdash yet independent\textemdash plant positions in a spatial way to identify single plants throughout all images of the whole growing season, even though they might not have been detected in every single image. Therefore, it is mandatory to align the image series in order to correct latent stitching errors or calibration glitches of the UAV's GPS sensor. Additionally, fixed georeference points to align the images to each other\textemdash as in the cauliflower dataset can be used. If the plants are seeded in certain line structures\textemdash for row crops, cereals is considerably more complex\textemdash one can do further steps in finding those lines and filtering \enquote{off-line} weed. This procedure is elaborated in the following.

\subsubsection{Aligning Plant Positions}\label{sec:align_plants}

\begin{figure}[t]
    \centering
    \includegraphics[width=\columnwidth]{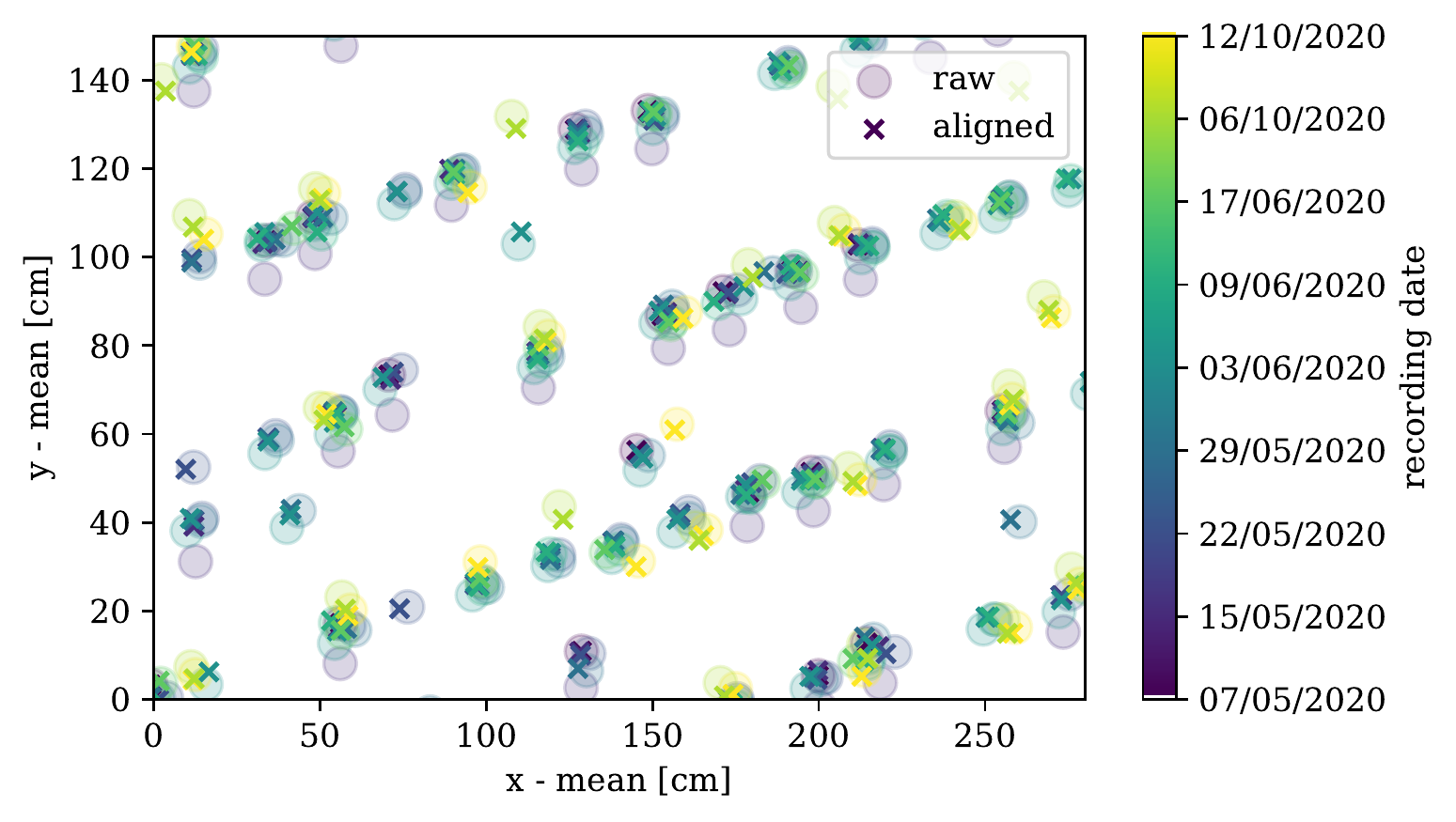}
    \caption{\textbf{Point cloud alignment.} Exemplary snippet from the application on peak positions. The color encodes the chronologically sorted acquisition date. Since the alignment procedure iterates over the layers with ascending cover ratio, this is not necessarily the alignment order. Shaded circles show the raw peak position data, whereas the crosses represent the same data after alignment. The data is already pre-aligned quite well so that only minimal improvements can be observed.}
    \label{fig:alignment}
\end{figure}

In order to avoid confusion of different plant IDs, the metric coordinates $\mathcal{P}(t)$ for each acquisition date $t$ are aligned to each other. Not every single plant is detected in each image. Therefore, the point clouds visualizing the plant positions are not congruent but still highly correlated. Additionally, also weed and other objects that can be observed in the VI images may be included in $\mathcal{P}(t)$. However, we can assume that these incorrectly detected objects are stationary in location as well and, thus, are helpful with the alignment. Afterwards, methods will be described to reduce unwanted objects from the detected peak positions. The main goal of this step is to avoid the largest GPS calibration errors. An application example is shown in Figure~\ref{fig:alignment}. In the following, the particular processing steps are described. Furthermore, a pseudocode algorithm for the position alignment can be found in the supplementary material.

Currently, $\mathcal{P}(t)$ contains (absolute) metric coordinates. However, for the next steps, it is convenient to centralize the coordinates by subtracting the centroid of all peak positions.
This results in the centralized coordinates
\begin{align}\label{eq:centralize_peaks}
    \begin{aligned}
        \mathcal{\bar{P}}(t) &= \Big\{\vec{x}_i(t) - \vec{x}_\text{mean}\Big\}_{i=0}^{N_p(t)}\,,\\
        \vec{x}_\text{mean}&\coloneqq\frac{1}{|\mathcal{T}^\ast|}\sum_{t\in\mathcal{T}^\ast}\frac{1}{N_p(t)}\sum_{i=1}^{N_p(t)}\vec{x}_i(t)\,.
    \end{aligned}
\end{align}
Generally, one can apply a transform containing constant shift, shearing, rotation, and scaling by an \textit{affine} transform given by the rule
\begin{equation}\label{eq:affine_transform}
    \mathbf{x'}(t) = \begin{pmatrix}
        T_{00}(t) & T_{01}(t) \\
        T_{10}(t) & T_{11}(t)
    \end{pmatrix}
    \mathbf{x}(t) +
    \begin{pmatrix}
        B_0(t) \\
        B_1(t) 
    \end{pmatrix}\,,
\end{equation}
where $\mathbf{x}(t)$ is the $(2\times N_p(t))$ matrix of the centralized plant positions $\mathcal{\bar{P}}(t)$. We assume that shearing effects are not relevant for the case of GPS calibration issues. By this assumption, one can concretize Equation~\eqref{eq:affine_transform} by only allowing shift, rotation, and scaling which results in
\begin{equation}\label{eq:rigid_transform}
    \mathbf{x'}(t) = S(t)\begin{pmatrix}
        \cos{\alpha(t)} & -\sin{\alpha(t)} \\
        \sin{\alpha(t)} &  \cos{\alpha(t)}
    \end{pmatrix}
    \mathbf{x}(t) +
    \begin{pmatrix}
        B_0(t) \\
        B_1(t) 
    \end{pmatrix}\,.
\end{equation}
Thus, we can reduce the generic transformation matrix elements $T_{ij}(t)$ to a scaling factor $S(t)$ and a rotation angle $\alpha(t)$. Initially, the images should be aligned at least roughly so that one can assume that $S(t)\approx 1$ and $\alpha(t)\approx\SI{0}{\degree}$. The shift parameters $B_0(t)$ and $B_1(t)$ should be in the order of several \si{\centi\meter}. We will perform what is usually called \textit{registration}. Constraining the transform to the setting in Equation~\eqref{eq:rigid_transform}, it is also referred to as \textit{rigid registration}. For the optimization, the method of \textit{coherent point drift} (CPD)~\cite{coherent_point_drift} is applied. Roughly speaking, it is based on a basis point set and a floating one, where the basis point set acts as data points. The floating-point set behaves as the centroids of a Gaussian mixture model (GMM). The objective is to minimize the negative log-likelihood function. At the minimum, the two point sets are considered to be optimally aligned\textemdash or registered\textemdash to each other. More detailed information is provided in~\cite{coherent_point_drift}.

In order to align all (centralized) \textit{point clouds} $\mathcal{\bar{P}}(t)$ corresponding to the acquisition dates, one has to apply the procedure multiple times for all layers in $\mathcal{T}^\ast$ step by step. Typically, the lower the cover ratio is, the more peaks are detected. One of the reasons is that the lower blurring bandwidth keep finer structures being visible for the low cover ratios. Therefore, it is useful to sort the acquisitions not by calendar date but by ascending cover ratio as estimated before\textemdash either by the growth function fit (cf.~\textit{\nameref{sec:growth_function}}), or by the cover ratio estimation (cf.~\textit{\nameref{sec:plant_seg}}). For rigid registration, we need a basis point set and a floating point set. As an initial basis set, the point cloud belonging to the image with the lowest cover ratio with acquisition date $t_\text{init}\in\mathcal{T}^\ast$ is chosen, i.e. $\mathcal{\bar{P}}(t_\text{init})$. The floating set is the point cloud with the next higher cover ratio. In order to make the alignment robust, only those points of the both sets are considered which have a next neighbor point in the other set within a maximum distance $d_\text{register}$.

After performing the registration, the resulting transformation (cf.~Equation \eqref{eq:rigid_transform}) is applied on all points of the floating layer. The new basis set is formed by grouping the aligned points that are in the vicinity of each other and calculate their centroid. The resulting set $\mathcal{\bar{P}}_\text{comb}$ contains only those group\textemdash or cluster\textemdash centroids. Points without neighbors inside a given threshold distance $d_\text{group}$ are considered as new groups. This distance should be chosen smaller than $d_\text{register}$. Subsequently, the method is repeated with the basis (centroid) set $\mathcal{\bar{P}}_\text{comb}$ and the point cloud with the next higher cover ratio as a floating set.

The developed grouping algorithm is applied in the final grouping as well, so we will make a detailed discussion later in section \textit{\nameref{sec:link_plants}}.

\subsubsection{Recognizing Seeding Lines}\label{sec:seeding_lines}

The peak positions are aligned to each other, which means that possible image acquisition errors should be mostly eliminated. Usually\textemdash like in the given datasets\textemdash the plants are seeded in straight, parallel seeding lines. Recognized peaks off those lines can then be considered as weed, other unwanted objects, or miscellaneous noise. Therefore, it is convenient to extract the seeding line positions in order to filter out those. In order to do this, we align the possible seeding lines with the $x$- or $y$-axis, since this is not necessarily the case for the GPS based images which are aligned to a geographical coordinate system. In this work, the seeding lines are aligned to be parallel to the $x$-axis. In contrast to approaches like in~\cite{intro_crop_weed_classification} that use the image information directly, we use the point cloud information to infer the seeding line positions and angles. Again, a pseudocode algorithm for seeding line recognition is given in the supplementary material.

Let us consider the acquisition dates $\mathcal{T}^\ast$ for which the peak point clouds are available. The idea of the seeding lines recognition is to apply a rotation transform on the joint point clouds $\bigcup_{t\in \mathcal{T}^\ast}\mathcal{\bar{P}}_\text{aligned}(t)=\mathcal{\bar{P}}_\text{aligned}$. The task is to find the right rotation angle $\alpha_s$. An established method in the field of computer vision to find regular structures like lines and circles in images is the \textit{Hough transform}~\cite{hough_transform}. It detects points in images that form line structures by transformation of those lines into a feature space (Hough space) consisting of the line angle $\theta$ and its minimum distance $d$ from origin. Thus, an infinite line in the (euclidean) image space is mapped to a point in the Hough space. The brute force approach of filling the Hough space is to scan in each non-zero image point a bunch of lines with different angles that intersect in this point. If another point is on this line, one increments the corresponding $(\theta, d)$-bin in Hough space. In the end, the infinite lines that are found in the image are visible by \enquote{nodes} in Hough space.

\begin{figure}[t]
    \centering
    \includegraphics[width=\columnwidth]{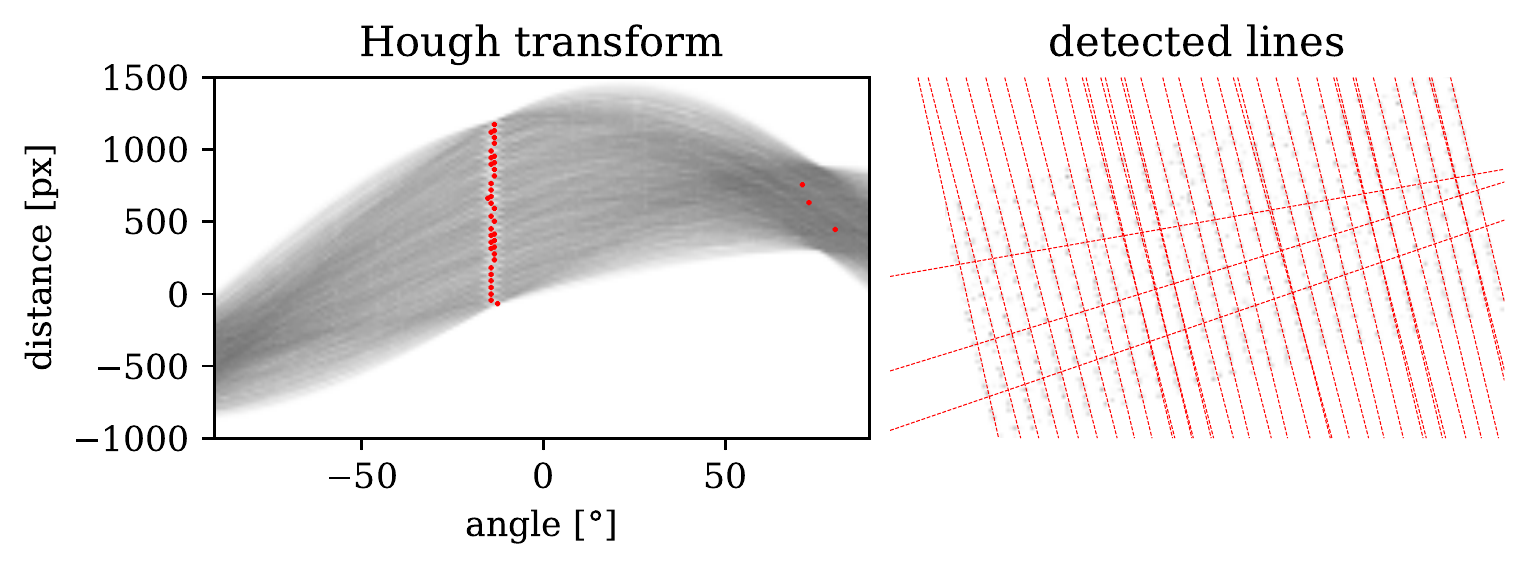}
    \caption{\textbf{Seeding line rotation angle determination with Hough transform.} The left plot shows the Hough transform applied to the rasterized image shown on the right. For reasons of visibility, the right image is rasterized with a 5 times larger bin width than for the actual Hough transformed image. One can see a bunch of nodes in the Hough transform at angles of roughly \SI{-14}{\degree}. Each node corresponds to one found line in the right image. By the vertical distance of the nodes, one can infer the line distances.}
    \label{fig:cropline_rot}
\end{figure}

To use this method, we first need to transfer $\mathcal{\bar{P}}_\text{aligned}$ into a rasterized image by binning the peaks into a 2D histogram with a fixed bin width. Secondly, we initially scan for a fixed amount of angles in $(-\SI{90}{\degree}, \SI{90}{\degree}]$. This results in a rather rough search of angles, although we expect that all seeding lines are rotated with a common angle. Other randomly found lines in the image can occur but they are at diffuse angles. The correct seeding line angles is expected to occur regularly so that we can expect a clustering around the correct seeding line angle. Thus, we use the principle of nested intervals: we divide the interval of queried angles into small bins. In the next step, we calculate the histogram of all found angles by using those bins. The diffuse angles will distribute over multiple bins, whereas our seeding line angle should accumulate in one or at least few bins. We take this bin, and maybe the surrounding bins as well, and take these as the new query interval which is then again divided into bins. After several iterations, one finds the common angle of all visible seeding lines $\alpha_s$. Figure~\ref{fig:cropline_rot} shows the Hough transform applied on a rasterized peak position image.

Since the Hough transform not only yields the line angles but also their distances, one might use this method directly to extract the seeding lines itself. However, it may happen that the Hough transform-based detection misses some less prominent line constellations. This is why we search the seeding line positions again in the point cloud, which is rotated by $\alpha_s$. Nevertheless, we can use the distance information of the Hough transform results as an estimate for the expected seeding line distances. To have a stable estimate which is robust against missed seeding lines, we use the median seeding line distance for our next step.

We consider only the $y$-coordinates $\vec{y}_s$ of the rotated point cloud. The principle is to count the points that are inside a certain window with a central $y$ coordinate and a size $\lambda$. A scan through all coordinates with a given precision $\rho$, i.e. the distance between two nearest window centers, results in the sum of included points against the window center position. Finally, one applies a peak finding method to extract the local maxima which represent the (sorted) seeding line positions $\vec{y}^\ast$. An example plot can be found in the supplementary material. The success of this method is strongly dependent on the right choice of $\lambda$ and $\rho$. In order to get this procedure to work stable, it is helpful to set $\lambda$ and $\rho$ in relation with the seeding line distance estimate of the Hough transform. The peak finding can also be tweaked considering, that one actually knows roughly where the peaks should be located.

\subsubsection{Filtering Weed}\label{sec:weed_filtering}

\begin{figure}[t]
    \centering
    \includegraphics[width=\columnwidth]{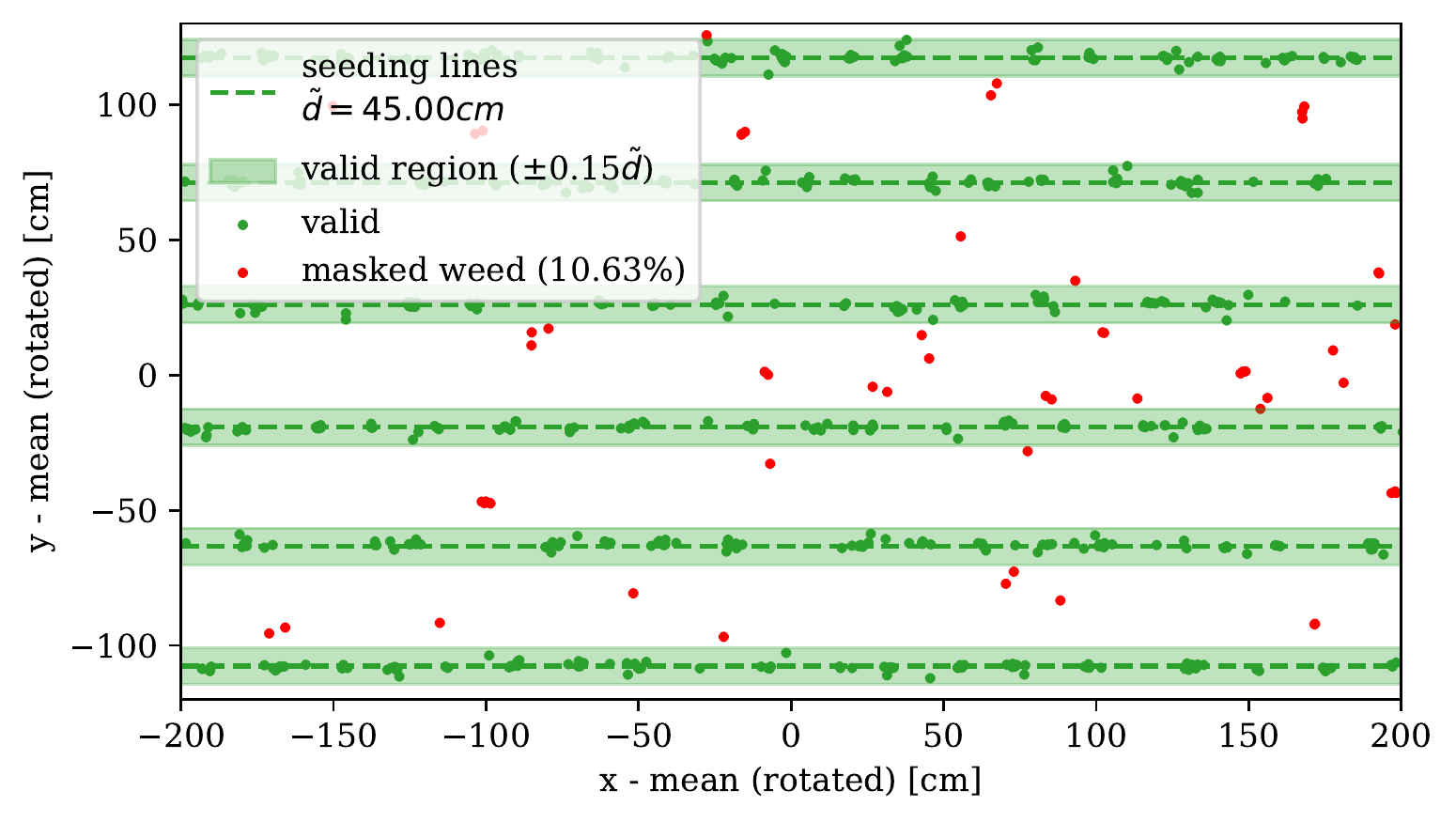}
    \caption{\textbf{Weed filtering.} Excerpt from peak positions where weed filtering is applied. Peaks inside the green shaded valid regions are considered to be valid plants (green dots) whereas peaks being outside are masked as weed (red dots).}
    \label{fig:weed_mask}
\end{figure}

One benefit of knowing the seeding line positions is that one can effectively filter out recognized objects that are located off the lines which are mainly weed or false detections. Usually, the seeding lines have a regular distance to their neighboring ones so that all distances are approximately by the mean distance. However, if the images contain few irregular seeding line distances, e.g. due to cart tracks in between, it is again helpful to prefer the median distance $\tilde{d}$ over the mean.
Additionally, we determine the distance of each point to its next seeding line by
\begin{equation}
    d_i = \min{|y_i\vec{1}-\vec{y}^\ast|}\,,
\end{equation}
where $y_i$ is the centralized, aligned, and rotated $y$-coordinate of the $i$-th plant position and $\vec{1}^\top = \{1\}^{\dim{\vec{y}^\ast}}$. Once all nearest distances $d_i$ are determined, a threshold factor $\vartheta_d$ can be set, that specifies, which proportion of the median distance $\tilde{d}$ should be the maximum distance of each plant position to their next seeding line to be considered as valid detection. Hence, the condition is
\begin{equation}
    d_i\leq\vartheta_d\tilde{d}\,.
\end{equation}
Figure~\ref{fig:weed_mask} shows an example weed filtering with threshold factor $\vartheta_d = \num{0.2}$.

\subsubsection{Linking Individual Plants During Time}\label{sec:link_plants}

After aligning and filtering the peak positions, we redefine $\mathcal{P}(t)$ to be the aligned, filtered plant positions and $\mathcal{P}\coloneqq\bigcup_{t\in\mathcal{T^\ast}}\mathcal{P}(t)$ to be the unified point cloud.

\begin{figure}[t]
    \centering
    \includegraphics[width=\columnwidth]{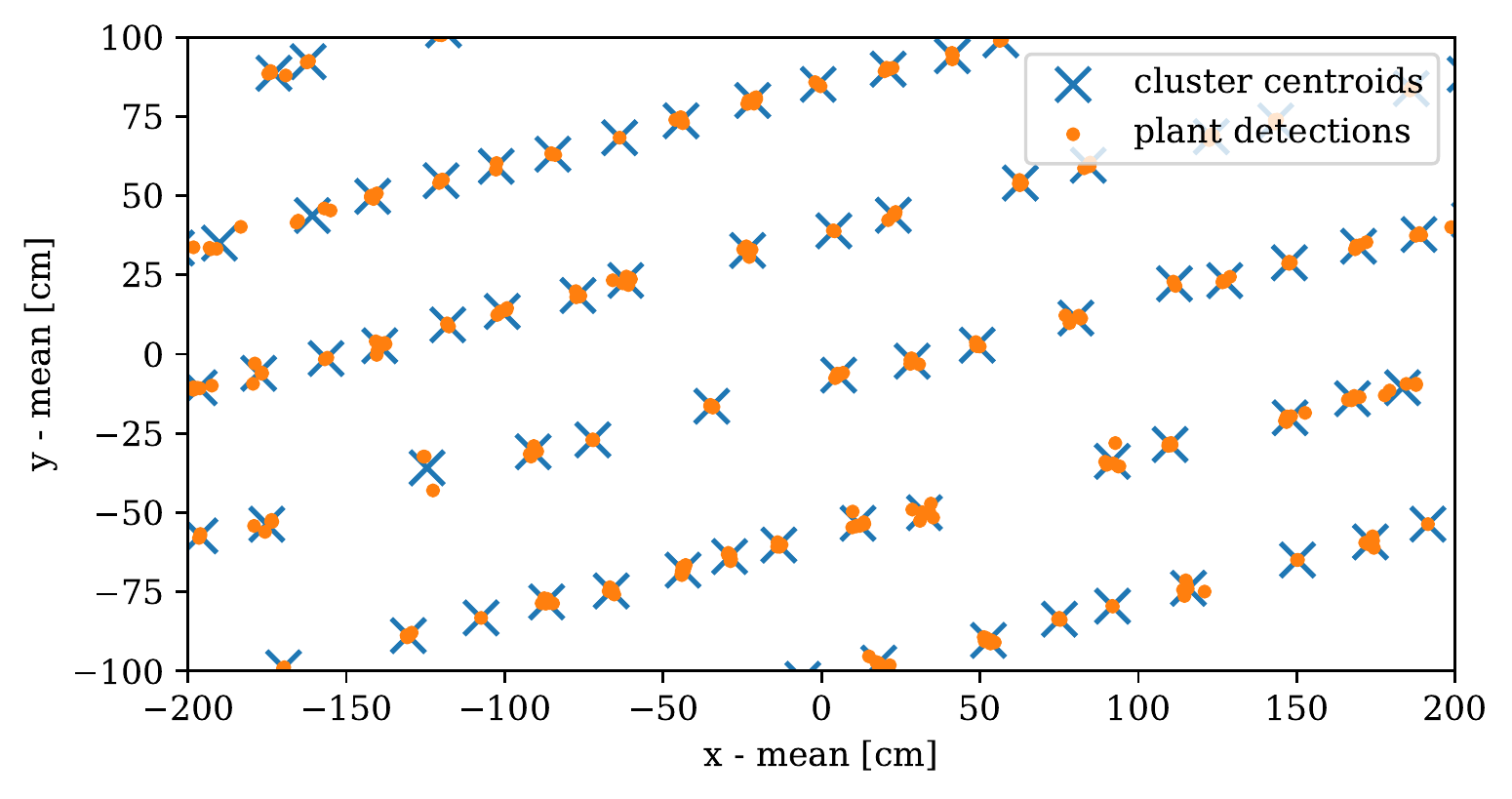}
    \caption{\textbf{Cluster centroids.} Each cluster represents a plant and incorporates the detections in the image series.}
    \label{fig:cluster_centroids}
\end{figure}

Furthermore, the corresponding image acquisition dates of all plants are summarized in the label vector $\vec{t}\in\mathcal{T^\ast}^{|\mathcal{P}|}$. The goal is to group the point cloud into clusters $\mathcal{G}\subset\mathcal{P}$ and thereby identify single plants recorded in those images where they are detected by the previous methods. An important coordinate of each cluster therefor is its centroid $\vec{\zeta}$ calculated by
\begin{equation}\label{eq:centroid}
    \vec{\zeta}_i = \frac{1}{|\mathcal{G}_i|} \sum_{\vec{x}\in\mathcal{G}_i} \vec{x}\,,
\end{equation}
where the $\vec{x}$ are the clusters' point coordinates in $\mathcal{P}$. We call the set of all $n_c$ cluster centroids $\mathcal{C}\coloneqq\{\vec{\zeta}_i\}_{i=0}^{n_c}$. Furthermore, we can postulate that each cluster contains not more than one member per acquisition date in $\mathcal{T}^\ast$. Using classical approaches like k-means or even more sophisticated ones like DBSCAN~\cite{dbscan} for the clustering, we can hardly make use this knowledge. Those common methods only handle single point clouds without integrating more information on those points. Thus, there may be multiple points from the same acquisition date in one cluster. To integrate the acquisition date information, we use an iterative\textemdash or \enquote{layer-wise}\textemdash approach comparable to a point-to-point registration, similar to the CPD method (cf.~\textit{\nameref{sec:align_plants}}). The different acquisition dates are considered as layers of points. Plant detections in plots with low cover ratio are more precise in general, so we start from the layer of the date with lowest cover ratio, as we did for the alignment. Each plant position is considered to be the centroid of a cluster containing\textemdash so far\textemdash one member. We label those by assigning a unique cluster ID for each centroid. Subsequently, the layer with the second lowest cover ratio is considered. We evaluate a Nearest Neighbor model based on the cluster centroids to get the euclidean distance for each point in the new layer to its next cluster centroid. Next, some decisions have to be made based on this next neighbor distance $d$. We set a maximum point-centroid distance $d_\text{max}$ for points to be considered as members of the corresponding clusters. Thus, if the new candidate is inside this distance ($d\leq d_\text{max}$), we label those points with the corresponding cluster ID. It can happen that multiple candidates are inside this distance. We then only register the closest point to the cluster and discard the other candidates. Points outside this distance ($d>d_\text{max}$), are assumed to be new clusters labeled new individual cluster IDs. Finally, the cluster centroids are recalculated by including the new candidates to their corresponding clusters. This procedure is repeated iteratively until all layers are processed. In the end, we get clusters with at most $|\mathcal{T}^\ast|$ members and a label vector $\vec{l}$ containing the corresponding cluster ID for each point. Figure~\ref{fig:cluster_centroids} shows an exemplary plot of the cluster centroids that now represent single plants. The IDs are ascending integer values starting from \num{0}. Discarded points get the label \num{-1}. The summarizing algorithm, written in pseudocode, can be found in the supplementary material.

\subsubsection{Further Filtering and Indirect Detections}\label{sec:further_filtering}

With the information of the cluster centroids $\mathcal{C}$ from the results of section \textit{\nameref{sec:link_plants}}, we can find a single plant also in images where it is not necessarily detected since the spatial location is fixed. But first, we start with a somehow \enquote{cosmetic} step. 

The label numbering in our clustering algorithm is continuous but\textemdash due to new clusters starting at different steps\textemdash not sorted. However, we may want to have the labels sorted for reasons of clarity and easier retrieval of the plants in in-field operations. A reasonable approach would be to use the seeding line coordinates $\vec{y}^*$ from section \textit{\nameref{sec:seeding_lines}} again in order to determine the seeding line ID for each plant group. The cluster centroids $\mathcal{C}$ are used to calculate the nearest seeding line and assign its ID $i_s$ to each cluster.
\begin{equation*}
    i_s = \arg{}\underset{y^\ast\in\vec{y}^\ast}{\min{}}|y_c-y^\ast|\,,
\end{equation*}
where $y_c$ is the $y$-coordinate of the cluster centroids.
Thus, we substitute the initial labels with new sorted labels: primarily sorted by the seeding line ID and secondarily by\textemdash in this case for instance\textemdash the $x$-coordinates of cluster centroids. A figure that shows this label sorting for a complete field is shown in the supplementary material.

Moreover, the special point in having the cluster centroid positions is that we can retrieve the plant positions in the high-cover-ratio images\textemdash those with acquisition dates $t\in \mathcal{T}\setminus\mathcal{T^\ast}$\textemdash where above methods could not be applied. Additionally, not every plant may be recognized in all analyzed images. In this context, we name the detections by the above methods to be \textit{direct detections}, whereas the reconstructed positions by using the cluster centroids are referred to as \textit{indirect detections}.
The aligning algorithm not only gives us the aligned coordinates of the (directly detected) plant positions, but also the transform vectors $\vec{R}(t)\coloneqq (S(t),\alpha(t),B_0(t),B_1(t))^\top$ for each image\footnote{... as long as the cover ratio is not too high. If it is, however, the alignment can not be performed and the transform vector is the identity transform $\vec{R}(t)\coloneqq(1,0,0,0)^\top$}. Thanks to the alignments being (rigid) affine transforms, they are invertible. Hence, it is
\begin{equation*}
    \mathbf{x}(t) =
    \frac{1}{S(t)}
    \begin{pmatrix}
        \cos{\alpha(t)} & \sin{\alpha(t)} \\
        -\sin{\alpha(t)} & \cos{\alpha(t)}
    \end{pmatrix}
    \left(
        \mathbf{x'}(t) -
        \begin{pmatrix}
            B_0(t) \\
            B_1(t) 
        \end{pmatrix} 
    \right)
\,.
\end{equation*}
by rearranging Equation~\eqref{eq:rigid_transform}. Thus, we can calculate the plant positions for indirect detections with this inverse transform in order to fill up the point clouds. In order to recover the directly detected plant positions in the individual images, we have to add the calibration error (cf.~\textit{\nameref{sec:align_plants}}) again. Please note further that $\mathbf{x}(t)$ are still the centralized UTM coordinates introduced in section \textit{\nameref{sec:align_plants}}. For a complete inverse transform into the GPS coordinate system and the rasterized image pixel system, respectively, we have to add the mean vector $\vec{x}_\text{mean}$ again. Finally, the indirect detections at each acquisition date are basically the detected cluster centroids inversely transformed into the coordinate systems of the image data at respective acquisition dates.

With all these above filtering rules, we have a labeled point set where each cluster has exactly $n_t=|\mathcal{T}|$ members, one for each acquisition date $t\in\mathcal{T}$. Finally, we claim the plants to be \enquote{cataloged}.    
   
\section{Evaluation and Discussion}\label{chap:results}

In this section, we evaluate our results on available ground truth information for both datasets. Furthermore, we introduce two conceivable exemplary application use cases for the workflow.

\subsection{Validation: Sugar Beet Dataset}\label{sec:results_cercospora}
With available manually annotated information on the plant positions as a reference, we can validate the above methods. For the sugar beet dataset, there is some ground truth data available which enables us to perform an evaluation of the binary classifiers of \textit{true positive} ($TP$), \textit{true negative} ($TN$), \textit{false positive} ($FP$), and \textit{false negative} ($FN$) detections. Since true negative detections make no sense in this application, measures like the accuracy are not appropriate. Therefore, we focus on two other measures called \textit{precision} and \textit{recall} defined by
\begin{equation*}
    \text{recall} = \frac{TP}{TP+FN}\,,\qquad \text{precision} = \frac{TP}{TP+FP}\,.
\end{equation*}
Descriptively spoken, the recall describes which ratio of really seeded plants our method catches whereas the precision measures the ratio of how many detections of our method indeed are real plants. We want to compare dot-like positions to each other. Thus, we additionally set a tolerance radius of maximum distance between a true plant position and its potential position of detection. We require the plants to be recognized with a maximum tolerance of \SI{8}{\centi\meter}, i.e. the maximum distance between detected plant position and ground truth annotation. The counting of $TP$, $FP$, and $FN$ is done thanks to a Nearest Neighbor approach. For each plant detection the ID and the distance to the next true plant position is evaluated. Then we iterate over the true plant positions. For each of them, we look at the distances from the plant detections that were assigned to those corresponding true plants. If there is at least one assignment inside our tolerance, we increment $TP$ by one and $FP$ by the number of remaining total assignments. If no assignment is inside the tolerance, we increment $FN$ by one and $FP$ by the number of total assignments. Since every recognized plant position is assigned to exactly one true position, we consider each true and recognized plant by this method. In the end, we obtain the binary classifier counts and can calculate precision and recall. Moreover, our method yield the plant positions by direct and indirect detections. In the ground truth data, the plant is not annotated if it is not visible, for instance in an early image. Since our method uses the inverse transformations of the plant position centroids $\mathcal{C}$, it would already find the plant even if it is not yet visible. Therefore, we ignore all \enquote{leading} indirect detections, so all that are earlier than the first direct detection in terms of acquisition date.

\begin{figure}[t]
    \centering
    \includegraphics[width=\columnwidth]{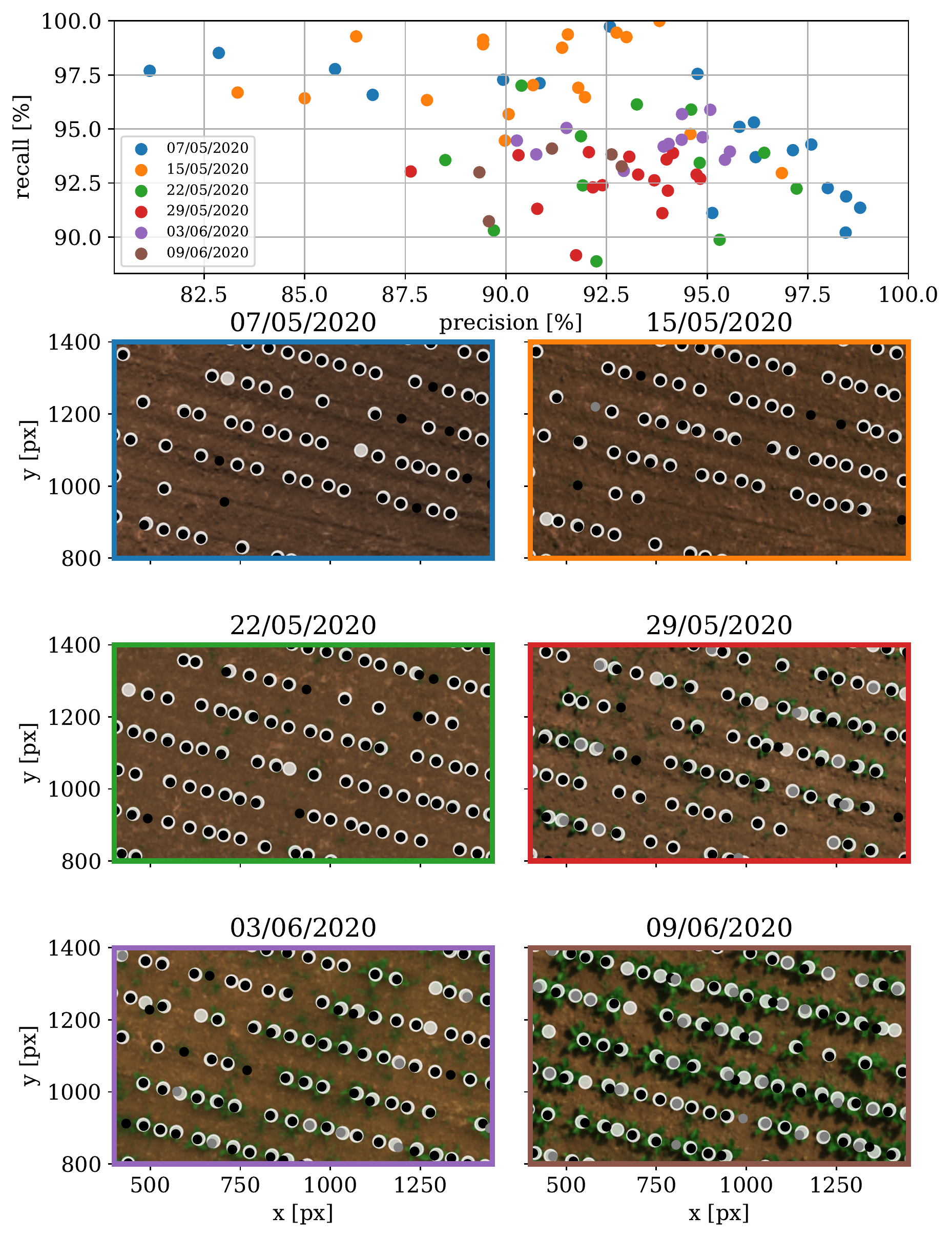}
    \caption{\textbf{Precision and recall evaluation summary for the sugar beet dataset.} The upper plot shows precision vs. recall for all field images with available ground truth data. Colors show the acquisition date of the corresponding image. Below, field excerpts are shown exemplarily\textemdash one for each available acquisition date. In these plots, the real plants are marked by white dots. Black dots represent direct detections via our peak detection method. The gray dots are indirect detections by using information of other images of the same field and back transform the positions in each image.}
    \label{fig:precision_recall_summary_cercospora}
\end{figure}

The upper plot of Figure~\ref{fig:precision_recall_summary_cercospora} shows evaluations of precision vs. recall for field images with available ground truth. Below, for each acquisition date (encoded by the color), an example image is shown. For most of the images, we achieve a precision of at least \SI{90}{\percent}. The recall is at least \SI{90}{\percent} for most of the images. As we expect, the plant detection gets slightly more inaccurate, the higher the cover ration is due to the lower punctiformity of the plants. However, the early plant detections at lower cover ratios are more confident. This should give the hint that high-cover-ratio images are inappropriate for plant detection and should only be used for the indirect retrieval of already known plant positions at the cluster centroids $\mathcal{C}$.

\subsection{Validation: Cauliflower Dataset}
\begin{figure}[t]
    \centering
    \includegraphics[width=\columnwidth]{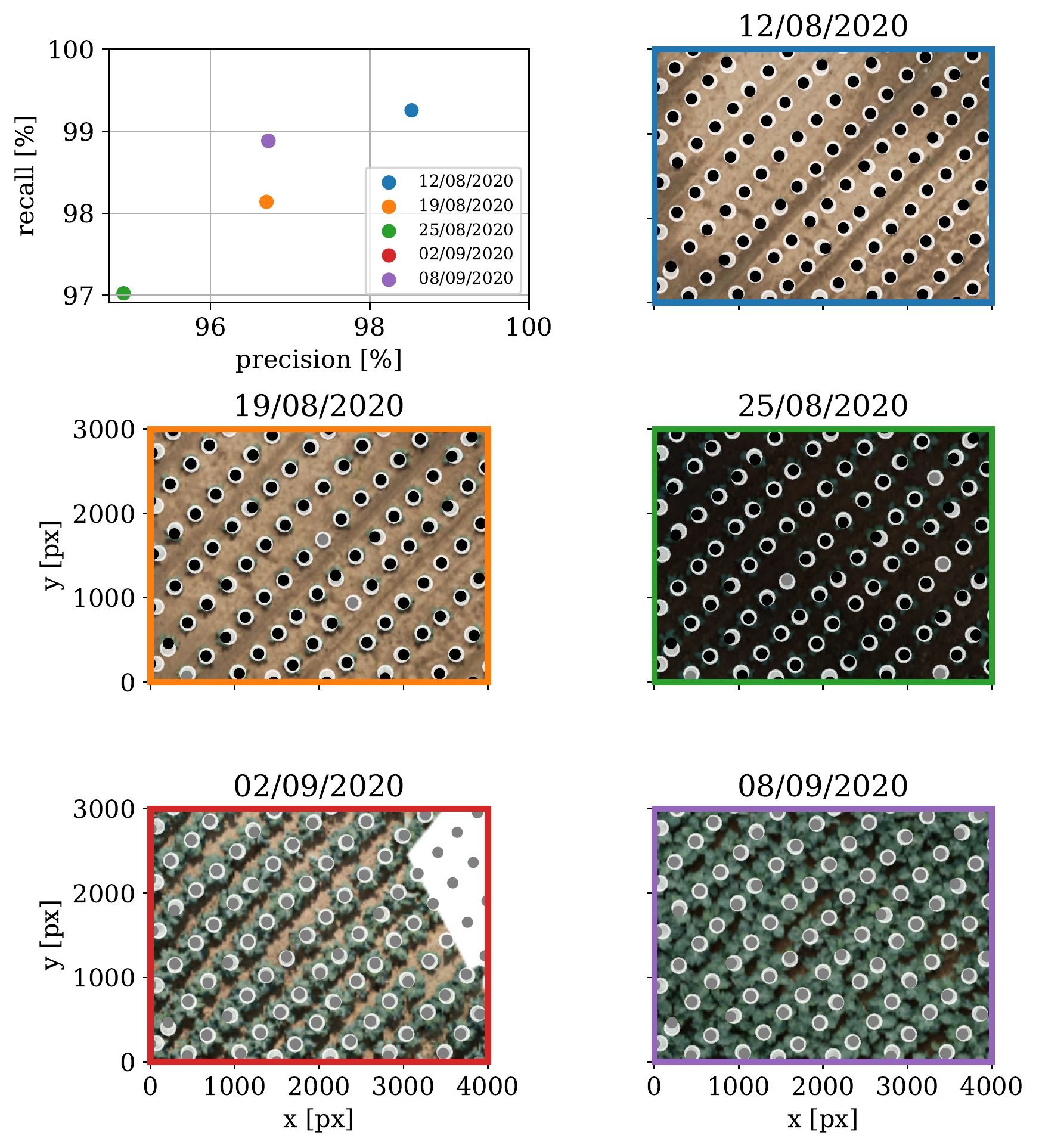}
    \caption{\textbf{Precision and recall evaluation summary for the Cauliflower dataset.} The upper plot shows precision vs.~recall for all field images with available reference detection data. Colors show the acquisition date of the corresponding image. Below, field excerpts are shown exemplarily\textemdash one for each available acquisition date. In these plots, the real plants are marked by white dots. Black dots represent direct detections via our peak detection method. The gray dots are indirect detections by using information of other images of the same field and back transform the positions in each image. As a reference detection method, plant positions are detected by a Mask R-CNN~\cite{mask_r-cnn}.}
    \label{fig:precision_recall_summary_cauliflower}
\end{figure}

We evaluate our method on the Cauliflower dataset analogously to the sugar beet dataset described in section \textit{\nameref{sec:results_cercospora}}. The difference is here that there is no \enquote{human annotated} ground truth data available. However, in the actual use case of the Cauliflower data, a Mask R-CNN~\cite{mask_r-cnn} trained with single Cauliflower plant images is used to extract the plant position. Our evaluation uses the Mask R-CNN results as reference for our own direct and indirect detections. The Mask R-CNN is applied on a single image (19/08/2020). Nevertheless, we consider the detections to be valid for all acquisition dates to evaluate our date-wise (direct and indirect) detections. As a maximum tolerance distance between our and the reference detections, we choose \SI{12}{\centi\meter} which is roughly the plant radius on the reference detection date. Five dates were available for the evaluation shown in Figure~\ref{fig:precision_recall_summary_cauliflower}. Due to our method, that\textemdash as long as being in a reasonable cover ratio range\textemdash the plants are detected in each image, our method yields more precise plant positions than the Mask R-CNN approach. This can be crucial for further image extraction of individual plants, where the plants should be centered in each extracted image. Our detections coincide well with the Mark R-CNN detections resulting in a precision above \SI{95}{\percent} and a recall above \SI{97}{\percent}. In particular, these benchmarks are better than for the sugar beet dataset. This is most likely due to the greater spacing and the resolution of the cauliflower plants compared to the sugar beets.

\subsection{Application: In-field Annotations}
\begin{figure}[t]
    \centering
    \includegraphics[width=0.7\columnwidth, trim={0 0 0 10cm}, clip]{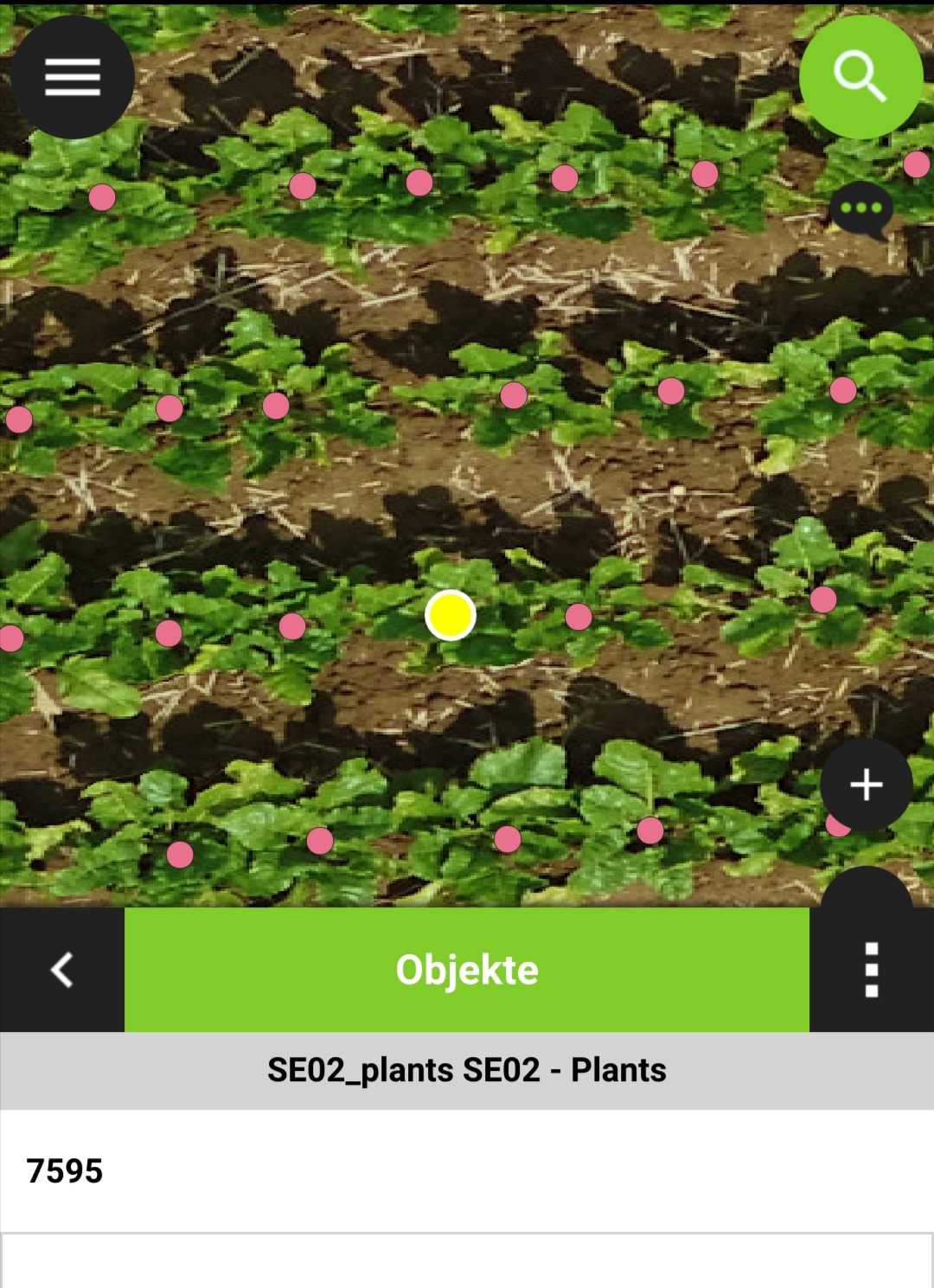}
    \caption{\textbf{Plant catalog for in-field observations with QField.} Exemplary screenshot on an Android smartphone. Detected plants can be accessed individually and various annotations like quality assessments, additional notes, or other images can be added.}
    \label{fig:qfield_example}
\end{figure}

We have all plant positions available GPS coordinates which enables processing the data with geographic information system applications like \textit{QGIS}~\cite{qgis}. For in-field operations like plant assessment or annotation, the plant catalog can support by providing a structured position sample of interesting plants. Additionally, there are some interesting applications like \textit{QField}~\cite{qfield} that are able to process georeferenced data directly on a smartphone. This is an excellent tool for farmers, who then can do in-field tasks directly \enquote{online} using the plant catalog and their current GPS position. Exporting the plant catalog for instance as KML file~\cite{kml} allows to enrich the georeferenced plant catalog with further annotation data like disease severity score, number of leaves, additional images, etc. Figure~\ref{fig:qfield_example} shows how this could possibly look like for an in-field scenario.

\subsection{Application: Image Extraction for Disease Severity Classification}

\begin{figure}[t]
    \centering
    \includegraphics[width=0.7\columnwidth]{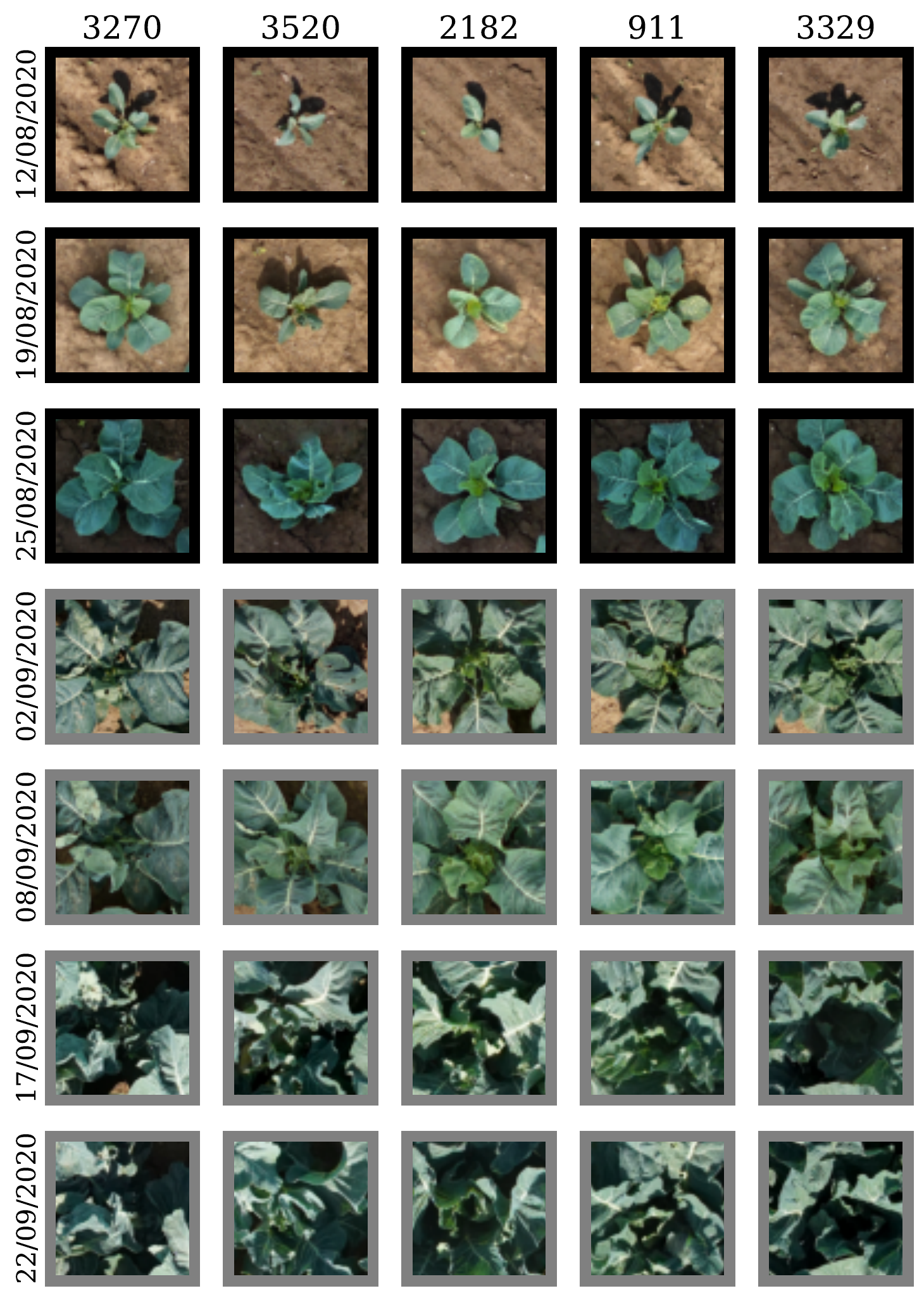}
    \caption{\textbf{Detected plant positions of cauliflower dataset.} 5 randomly picked image series of plant RGB images detected by our method. The acquisition date increases downwards. Black frames annotate direct, gray frames indirect detections.}
    \label{fig:cauliflower_images_overview}
\end{figure}

An interesting use case of the gathered plant catalog is image extraction. The catalog makes it possible to accumulate a time series of images of individual plants. By inverse transformation of the cluster member coordinates in their respective image pixel-based coordinate system (cf.~\textit{\nameref{sec:further_filtering}}) one gets the, say, \enquote{original} pixel positions of the plants. Next, we simply define a frame around those pixels and get smaller \enquote{tiles}. These can then be used for further steps, for instance as a training set for Neural Network architectures. The fact that the images are linked to individual plants at multiple dates enables this dataset to be used not only for spatially related but also for time series analyses. Figure~\ref{fig:cauliflower_images_overview} shows a few example image tiles for the cauliflower dataset. Other example image tiles for the sugar beet dataset can be found in the supplementary material.

\begin{figure}[t]
    \centering
    \includegraphics[width=0.8\columnwidth]{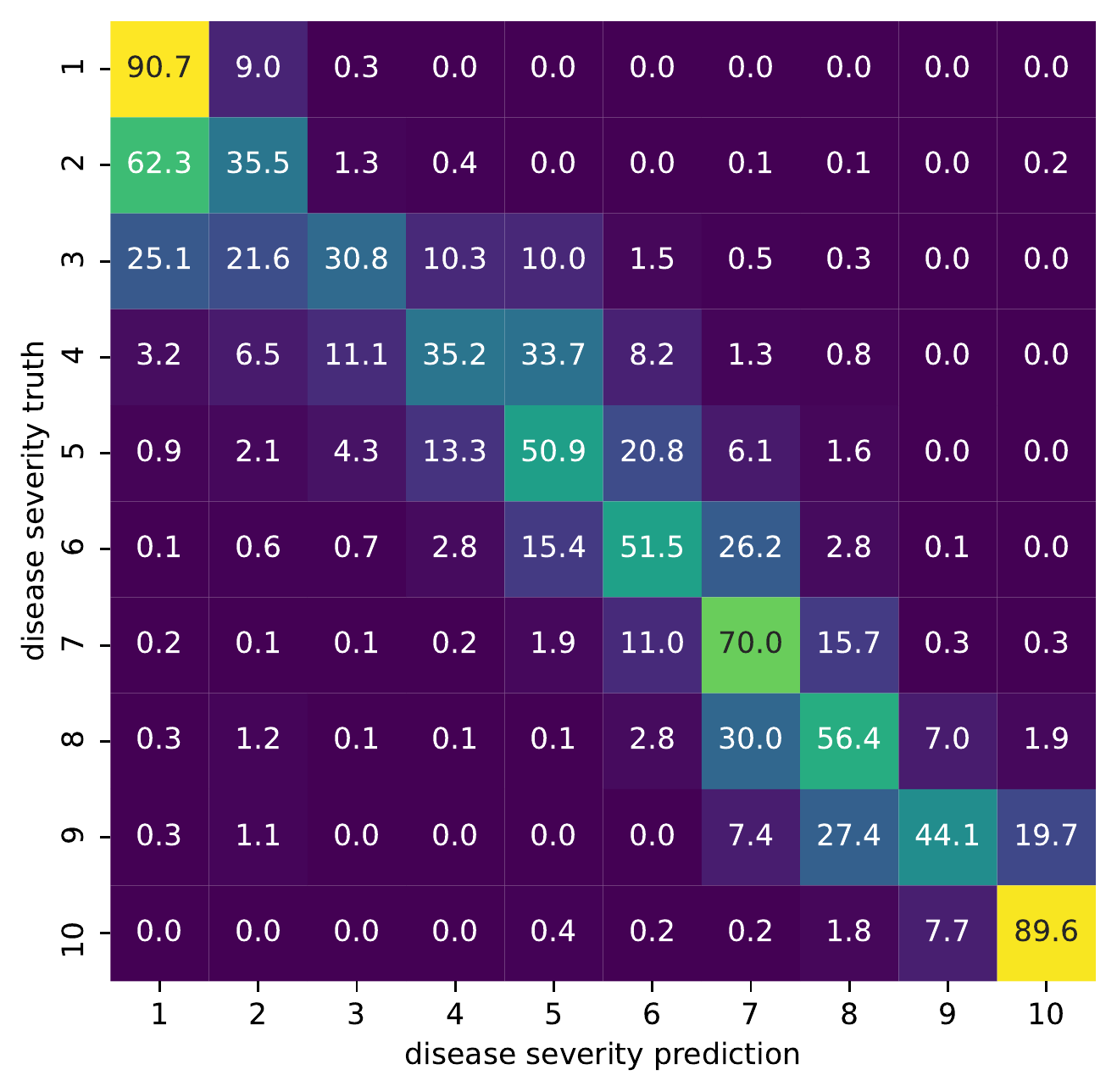}
    \caption{\textbf{Confusion matrix for the disease severity classification example.} The values inside the heatmap are percentages of the corresponding category. Percentages are normalized to the truth, so that all rows sum up to \SI{100}{\percent}.}
    \label{fig:kws_confusion_matrix}
\end{figure}

As a brief sample application, we can use the extracted images to automatize the plant health rating by image classification. The rating scale\footnote{german: Bonitur} is based on a scale by the corporation \textit{KWS SAAT SE \& Co. KGa} (KWS)~\cite{cercospora_kws,cercospora_kws2}. Five plant health states are shown and ordered into classes 1, 3, 5, and 9. We complement classes in between, resulting in 10 disease severity degrees from 1 (completely healthy) to 10 (completely diseased). A subsample of about \num{4000} extracted plant images was annotated to train a Convolutional Neural Network (CNN). In order to increase the size of the training sample, we augment the data by sampling further images of the same plants with random shift and rotation. Further augmentation methods like sampling different light conditions are conceivable as well.~\cite{data_augmentation} For the example CNN, we use a standard ResNet-50 model~\cite{resnet} which is pretrained on the ImageNet dataset~\cite{imagenet} consisting of RGB images. We adapt the network to our use case (5 channels, 10 output classes) by adding two convolutional layers before the ResNet-50 network successively reducing the 5 channels to 3. After the ResNet-50 layers, we add a dense layer reducing the 2048 classes to our 10 disease severity degrees. The confusion matrix between true and predicted disease severity degrees in Figure~\ref{fig:kws_confusion_matrix} shows that the classification is largely possible. However, the discrimination between the first 5 degrees seems to be challenging. Certainly, further research can be done, since this is beyond the scope of this work.

\section{Conclusion}

The presented workflow shows a very satisfactory detection performance for both considered datasets. By choosing RGB-only VIs for the peak detection, we demonstrated successfully, that for our plant cataloging, RGB information is sufficient. This could be important for use cases\textemdash as shown in the cauliflower dataset\textemdash where a multi- or even hyperspectral data acquisition is not feasible for whatever reason. Nevertheless, we may not exclude that the peak detection performance could be improved having available data beyond the optical spectrum. The high degree of automatization enables the analysis of large-scale data. Particularly for those consisting of multiple smaller fields, like in the sugar beet dataset, many workflow steps can be processed in parallel.
High-level deep learning models that are developed recently require large amounts of data, especially for tasks involving image processing like, e.g., classification and instance segmentation with Mask R-CNNs. In principle, non-invasive remote sensing and UAVs enable the application to large-scale agricultural experiments and corresponding data. By automatizing the plant cataloging and providing a data framework, our work helps to exploit the full potential of UAV imaging in agricultural contexts.

\section{Abbreviations}
CPD: Coherent Point Drift; CLS: Cercospora Leaf Spot disease; CNN: Convolutional Neural Network; GCP: Ground Control Point; GIS: Geographic Information System (GIS); GLI: Green Leaf Index; GLONASS: Global Navigation Satellite System; GPS: Global Positioning System; LiDAR: Light Detection And Ranging; NGRDI: Normalized Green/Red Difference Index; OSAVI: Optimized Soil Adjusted Vegetation Index; RTK: Real-time Kinematic Positioning; UAV: Unmanned Aerial Vehicle; UGV: Unmanned Ground Vehicle; UTM: Universal Transverse Mercator Coordinate System; VI: Vegetation Index

\section{Competing Interests}
The authors declare that they have no competing interests.

\section{Funding}
This work has been funded by the Deutsche Forschungsgemeinschaft (DFG, German Research Foundation) under Germany’s Excellence Strategy – EXC 2070 – 390732324 and partially by the European Agriculture Fund for Rural Development with contribution from North-Rhine Westphalia (17-02.12.01 - 10/16 – EP-0004617925-19-001).
The work is partly funded by the German Federal Ministry of Education and Research (BMBF) in the framework of the international future AI lab "AI4EO -- Artificial Intelligence for Earth Observation: Reasoning, Uncertainties, Ethics and Beyond" (Grant number: 01DD20001).

\section{Authors' Contributions}
The workflow was designed and implemented by MG. The sugar beet experiments were carried out by FI as well as preprocessing and annotation of the corresponding data. The Cauliflower dataset was provided and preprocessed by JK. MG, FI, and JK drafted the manuscript. CB, AKM, and RR supervised the research. All authors approved the final manuscript.

\section{Availability of Source Code}
The source code of our workflow is available in the following repository:
\begin{itemize}
\item Project name: Plant Extraction Workflow
\item GitHub repository: \url{https://github.com/mrcgndr/plant_extraction_workflow}
\item Operating system(s): Platform independent
\item Programming language: Python (3.9 or higher)
\item License: Apache License 2.0
\item Other requirements: None
\end{itemize}

\bibliographystyle{unsrt}  
\bibliography{references}  

\clearpage

\appendix
\section{Supplementary Material}

\subsection{Algorithms}

Algorithms \ref{alg:align_plants}, \ref{alg:seeding_line_recognition}, and \ref{alg:cluster_points} show some important steps of our workflow written in pseudocode.

\begin{algorithm*}[h]
    \caption{Plant position alignment}
    \begin{algorithmic}[1]
        \footnotesize
        \Require
            \Statex $\mathcal{\bar{P}}(t)$: centralized point clouds
                \Comment{\parbox[t]{0.5\linewidth}{eq. \eqref{eq:centralize_peaks}}}
            \Statex $\mathcal{T}^\ast$: acquisition dates sorted by cover ratio in ascending order
            \Statex $t_\text{init}\in\mathcal{T}^\ast$: acquisition date of initial point cloud with the lowest cover ratio
            \Statex $d_\text{register}$: maximum next neighbor distance between basis and floating layer
            \Statex $d_\text{group} < d_\text{register}$: maximum distance from centroid for potential group members
        \Ensure
            \Statex $\mathcal{\bar{P}}_\text{aligned}(t)$: aligned centralized point clouds
            \Statex $\vec{R}(t) \coloneqq (S(t), \alpha(t), B_0(t), B_1(t))^\top$: transforms for each layer
                \Comment{\parbox[t]{0.5\linewidth}{eq. \eqref{eq:rigid_transform}}}
    \State initialize $\vec{R}(t_\text{init}) \gets (1, 0, 0, 0)^\top$
    \State initialize $\mathcal{\bar{P}}_\text{aligned}(t_\text{init}) \gets \mathcal{\bar{P}}(t_\text{init})$
    \State initialize $\mathcal{\bar{P}}_\text{comb} \gets \mathcal{\bar{P}}(t_\text{init})$
        \Comment{\parbox[t]{0.5\linewidth}{point cloud of cluster centroids (initialized by first layer)}}
    \For{$t \in \mathcal{T}^\ast\setminus \{t_\text{init}\}$}
        \State $\texttt{nn\_model\_comb} \gets \texttt{nearestNeighborModel}(\mathcal{\bar{P}}_\text{comb})$
            \Comment{\parbox[t]{0.5\linewidth}{fit Nearest Neighbor model to grouped centroid layer}}
        \State $\vec{d}_\text{new} \gets \texttt{nn\_model\_comb.evaluate}(\mathcal{\bar{P}}(t))$
            \Comment{\parbox[t]{0.5\linewidth}{get distances for all points of new layer to their next neighboring centroid}}    

        \State $\texttt{nn\_model\_new} \gets \texttt{nearestNeighborModel}(\mathcal{\bar{P}}(t))$
            \Comment{\parbox[t]{0.5\linewidth}{fit Nearest Neighbor model to new layer}}
        \State $\vec{d}_\text{comb} \gets \texttt{nn\_model\_new.evaluate}(\mathcal{\bar{P}}_\text{comb})$
            \Comment{\parbox[t]{0.5\linewidth}{get distances for all centroids to their next neighboring point of the new layer}}   

        \State $\vec{R}(t) \gets \texttt{rigidCPD}(\{\mathcal{\bar{P}}_\text{comb}\,|\,d_\text{comb}\leq d_\text{register}\}, \{\mathcal{\bar{P}}(t)\,|\,d_\text{new}\leq d_\text{register}\})$
            \Comment{\parbox[t]{0.5\linewidth}{perform rigid CPD registration with subsets}}
        \State $\mathcal{\bar{P}}_\text{aligned}(t) \gets \texttt{transform}(\mathcal{\bar{P}}(t), \vec{R}(t))$
            \Comment{\parbox[t]{0.5\linewidth}{transform complete set using eq. \eqref{eq:rigid_transform}}}
        \State $\mathcal{\bar{P}}_\text{comb} \gets \texttt{clusterPoints}(\mathcal{\bar{P}}_\text{comb} \cup \mathcal{\bar{P}}_\text{aligned}(t), d_\text{group})$
            \Comment{\parbox[t]{0.5\linewidth}{get cluster centroids (cf. Algorithm~\ref{alg:cluster_points})}}
    \EndFor
  \end{algorithmic}
  \label{alg:align_plants}
\end{algorithm*}

\begin{algorithm*}[h]
    \caption{Seeding line recognition}
    \begin{algorithmic}[1]
        \footnotesize
        \Require
            \Statex $\mathcal{T}^\ast$: acquisition dates with given peak positions
            \Statex $\mathcal{Q}\coloneqq \bigcup_{t\in \mathcal{T}^\ast}\mathcal{\bar{P}}_\text{aligned}(t)$: centralized aligned point clouds
            \Statex $\mathbf{q}\in\mathbb{R}^{2\times n},\,n=|\mathcal{Q}|$: matrix of point coordinates in point cloud $\mathcal{Q}$ with $\vec{q}_i = (x_i, y_i)^\top$ 
            \Statex $n_b$: number of bins for Hough angles histogram
            \Statex $n_+$: number of surrounding bins to respect for nesting iteration
            \Statex $\lambda$: window length for seeding line position scan
            \Statex $\rho$: precision for seeding line position scan
        \Ensure
            \Statex $\vec{y}^\ast$: seeding line positions
            \Statex $\alpha_s$: rotation angle
            
    \State initialize angle interval $[\alpha_\text{min}, \alpha_\text{max}] \gets [-\SI{90}{\degree}, \SI{90}{\degree}]$
    \State $\mathbf{I} \gets \texttt{2d\_histogram(}\mathbf{q}\texttt{)}$
        \Comment{\parbox[t]{0.6\linewidth}{raster point coordinates into image with given bin width}}

    \Repeat
        \State $\mathbf{H} \gets \texttt{hough\_transform}( \mathbf{I}, [\alpha_\text{min}, \alpha_\text{max}])$
            \Comment{\parbox[t]{0.6\linewidth}{perform Hough transform on image in given interval of angles}}
        \State $\vec{d}, \vec{\alpha} \gets \texttt{find\_lines}(\mathbf{H})$
            \Comment{\parbox[t]{0.6\linewidth}{find nodes in Hough image yielding line distances $\vec{d}$ and corresponding angles $\vec{\alpha}$}}
        
        \State $\mathcal{B} \gets$ set of $n_b$ equal binned intervals in $[\alpha_\text{min}, \alpha_\text{max}]$
        \State $\vec{h} \gets \texttt{histogram}(\vec{\alpha},\mathcal{B})$ 
            \Comment{\parbox[t]{0.6\linewidth}{$\vec{h}$ counts the number of nodes in the corresponding bin in $\mathcal{B}$}}
        \State $\mu = \frac{1}{\dim \vec{\alpha}}\sum_{i=1}^{\dim\vec{\alpha}}\alpha_i$
            \Comment{\parbox[t]{0.6\linewidth}{take the interval, where most of the nodes are included}}
        \State $[\alpha_\text{min},\alpha_\text{max}] \gets \bigcup_{i=\max\{1;\arg\max{\vec{h}}-n_+\} }^{\min\{n_b;\arg\max{\vec{h}}+n_+\}} \mathcal{B}_i$
    \Until{$\mu = \alpha_i \forall \alpha_i \in \vec{\alpha}$}
    
    \State $\alpha_s \gets \alpha_\text{min}$

    \State $\mathbf{q_s} \gets \begin{pmatrix}
            \cos{\alpha_s} & -\sin{\alpha_s} \\
            \sin{\alpha_s} &  \cos{\alpha_s}
            \end{pmatrix}\mathbf{q}$
        \Comment{\parbox[t]{0.6\linewidth}{rotated point cloud defined analoguously to $\mathbf{Q}$, thus $\vec{q}_{s,i}=(x_{s,i}, y_{s,i})^\top$}}
    \State initialize point sum vector $\vec{\sigma}\gets\vec{0}$
    \State initialize $i\gets 0$
    \State initialize $y_\text{test}\gets\min\vec{y}_s-\lambda$
    \While{$y_\text{test} < \max\vec{y}_s+\lambda$}
        \State $\sigma_i\gets \sum_{l=1}^{\dim \vec{y}_s} I_{\{y_\text{test}-\lfloor\frac{\lambda}{2}\rfloor\leq y_{s,l}< y_\text{test}+\lfloor\frac{\lambda}{2}\rfloor\}}$
        \State $y_\text{test}\gets y_\text{test}+\rho$
        \State $i\gets i+1$
    \EndWhile
    \State $\vec{p}_\text{peaks}\gets\texttt{peakfinder}(\vec{\sigma})$
        \Comment{\parbox[t]{0.6\linewidth}{finds local maximum peak positions representing seed line positions}}
    \State $\vec{y}^\ast\gets (\min\vec{y}_s-\lambda)\vec{1} + \rho\vec{p}_\text{peaks}$
        \Comment{\parbox[t]{0.6\linewidth}{$\vec{1}\coloneqq(1,\dots,1)^\top$}}
    \end{algorithmic}
    \label{alg:seeding_line_recognition}
\end{algorithm*}

\begin{algorithm*}[h]
    \caption{Iterative point clustering}
    \begin{algorithmic}[1]
        \footnotesize
        \Require
            \Statex $\mathcal{P}$: aligned point clouds
            \Statex $\mathcal{T}^\ast$: acquisition dates with plant position information sorted by cover ratio in ascending order
            \Statex $t_\text{init}\in\mathcal{T}^\ast$: acquisition date of initial point cloud with the lowest cover ratio
            \Statex $d_\text{max}$: maximum distance from centroid for potential cluster members
        \Ensure
            \Statex $\vec{l}$: label vector of cluster ID for each point
            \Statex $\mathcal{C}$: cluster centroids
    \State initialize $\vec{l} \gets \{-1\}^{|\mathcal{P}|}$
        \Comment{\parbox[t]{0.6\linewidth}{-1 = not assigned to any cluster}}
    \State $\{\vec{l}\,|\,t=t_\text{init}\} \gets \{0,1,\dots\}$
        \Comment{\parbox[t]{0.6\linewidth}{assign individual labels for each point of first layer}}
    \State initialize $\mathcal{C} \gets \{\mathcal{P}\,|\,t=t_\text{init}\}$
    \For{$t' \in \mathcal{T}^\ast\setminus \{t_\text{init}\}$}
        \State $\texttt{nn\_model} \gets \texttt{nearestNeighborModel}(\mathcal{C})$
            \Comment{\parbox[t]{0.6\linewidth}{fit Nearest Neighbor model}}
        \State $\vec{d}, \vec{\Tilde{l}} \gets \texttt{nn\_model.evaluate}(\{\mathcal{P}\,|\,t=t'\})$
            \Comment{\parbox[t]{0.6\linewidth}{get next neighbor distance and ID for all points of respective layer; next neighbor ID corresponds to cluster ID}}
        \State initialize $\vec{m} \gets \{1\}^{\dim{\vec{d}}}$
            \Comment{\parbox[t]{0.6\linewidth}{vector for new cluster member candidates\\1 = valid member candidate for existing cluster\\0 = valid member candidate for new cluster\\-1 = excluded due to multiple candidates for a single cluster}}
        \State $\{ m\in\vec{m}\,|\,d>d_\text{max}\}\gets 0$
            \Comment{\parbox[t]{0.6\linewidth}{assign points with no existing cluster in their vicinity to a new cluster ($m\gets 0$)}}
        \For{$\Tilde{l}'\in\{\vec{\Tilde{l}}\}$}
            \Comment{\parbox[t]{0.6\linewidth}{iterate over the set of cluster IDs in $\vec{\Tilde{l}}$}}
            \State $\Big\{m\in\vec{m}\,\Big|\,d\neq \min\{d\in\vec{d}\,|\,\Tilde{l}=\Tilde{l}'\}\Big\} \gets -1$
                \Comment{\parbox[t]{0.6\linewidth}{for multi-assignments: get candidates for one cluster and keep the one with the minimal cluster distance; exclude the others ($m\gets -1$)}}
        \EndFor
        \State $\{l\in\vec{l}\,|\,t=t'\land m=1\} \gets \{\Tilde{l}\in\vec{\Tilde{l}}\,|\,m=1\}$
            \Comment{\parbox[t]{0.6\linewidth}{assign valid member candidates for existing clusters to respective ones}}
        \State $\{l\in\vec{l}\,|\,t=t'\land m=0\} \gets \{\max({\vec{l}})+1, \max({\vec{l}})+2, \dots\}$
            \Comment{\parbox[t]{0.6\linewidth}{give new cluster labels to valid member candidates for new clusters}}
        \State $\mathcal{C}\gets\emptyset$
            \Comment{\parbox[t]{0.6\linewidth}{reset clusters before recalculation}}
        \For{$l'\in\{\vec{l}\}\setminus \{-1\}$}
            \Comment{\parbox[t]{0.6\linewidth}{iterate over the set of valid cluster IDs in $\vec{l}$}}
            \State $\mathcal{G}\gets\{\mathcal{P}\,|\,l=l'\}$
                \Comment{\parbox[t]{0.6\linewidth}{consider single cluster with cluster ID $l'$}}
            \State $\mathcal{C}\gets\mathcal{C}\cup\Big\{\frac{1}{|\mathcal{G}|}\sum_{\vec{x}\in\mathcal{G}}\vec{x}\Big\}$
                \Comment{\parbox[t]{0.6\linewidth}{recalculate cluster centroids with new cluster assignments (cf. eq.~\eqref{eq:centroid})}}
        \EndFor
    \EndFor
  \end{algorithmic}
  \label{alg:cluster_points}
\end{algorithm*}

\subsection{Additional plots}

\begin{figure}[h]
    \centering
    \includegraphics[width=\columnwidth]{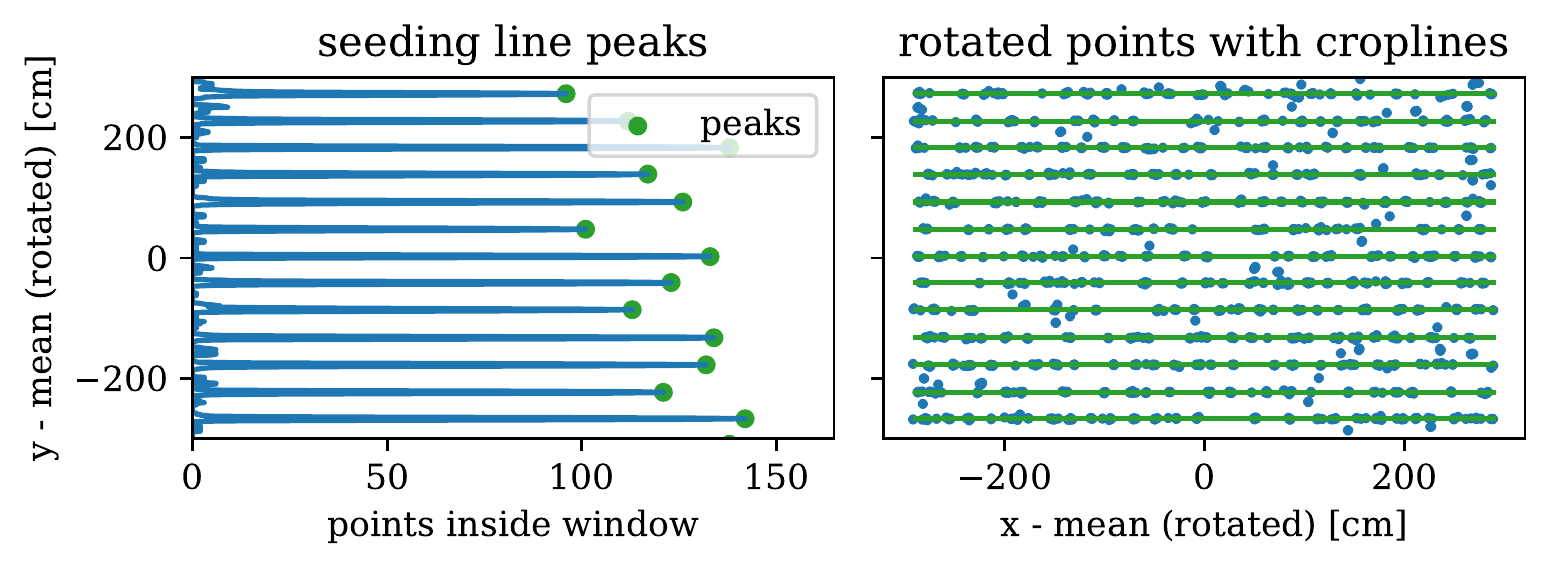}
    \caption{\textbf{Seeding line recognition.} For the seeding line recognition, the points inside a window with the size $\lambda=\SI{32}{px}$ are scanned with a precision of $\rho=\SI{0.5}{px}$ as seen in the left plot. In this example, 29 valid peaks are found by k-means. They represent the center $y$-coordinates of the seeding lines shown in the right plot with the corresponding (rotated) peak positions.}
    \label{fig:cropline_peaks}
\end{figure}

The seeding line recognition based on counting the number of points (plant detections) inside a moving window is shown in Figure~\ref{fig:cropline_peaks}.

\begin{figure}[h]
    \centering
    \includegraphics[width=\columnwidth]{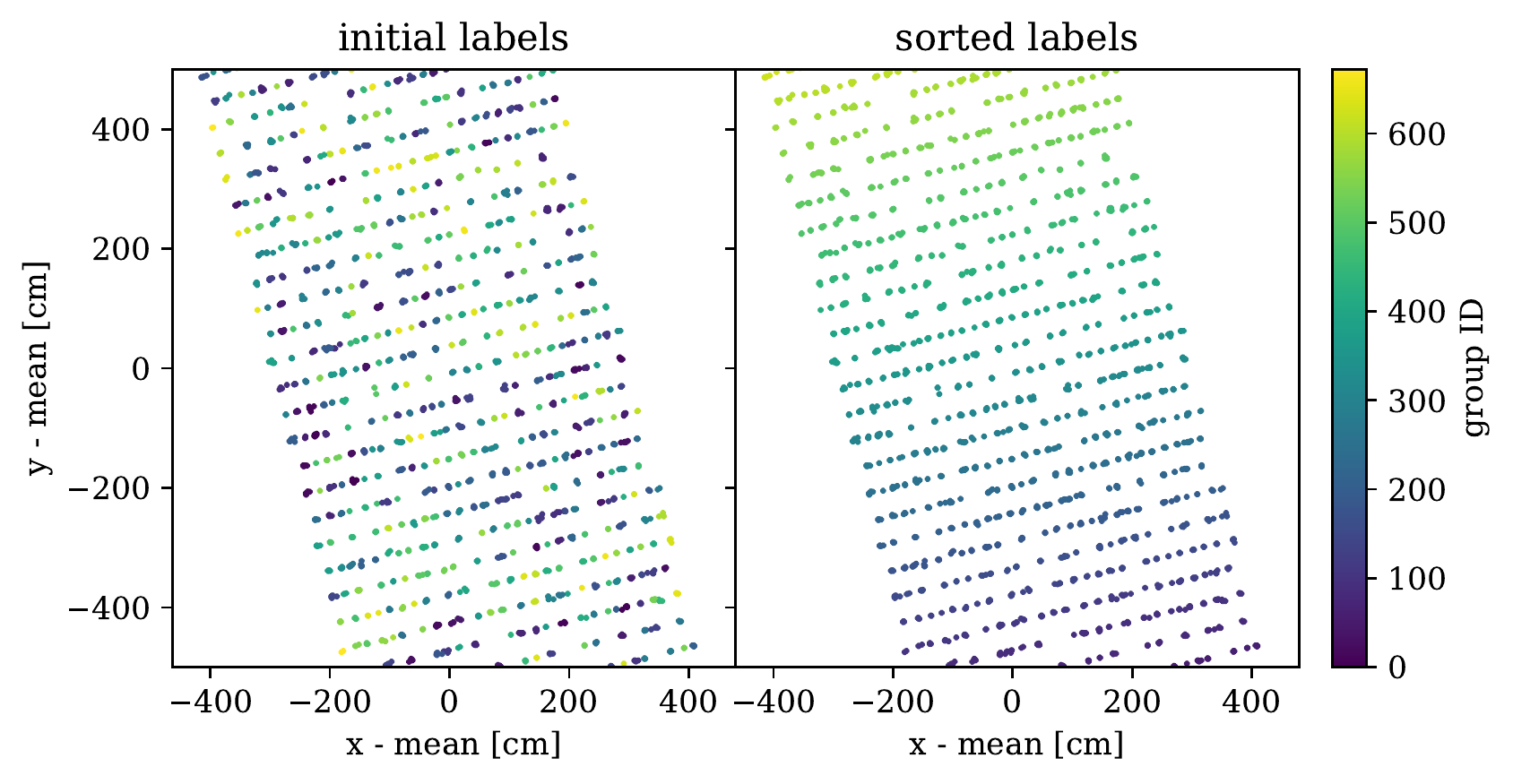}
    \caption{\textbf{Group label sorting.}}
    \label{fig:sorted_labels}
\end{figure}

The label sorting that is done in section \textit{\nameref{sec:further_filtering}} is shown in Figure~\ref{fig:sorted_labels}.

\begin{figure}[h]
    \centering
    \includegraphics[width=0.6\columnwidth]{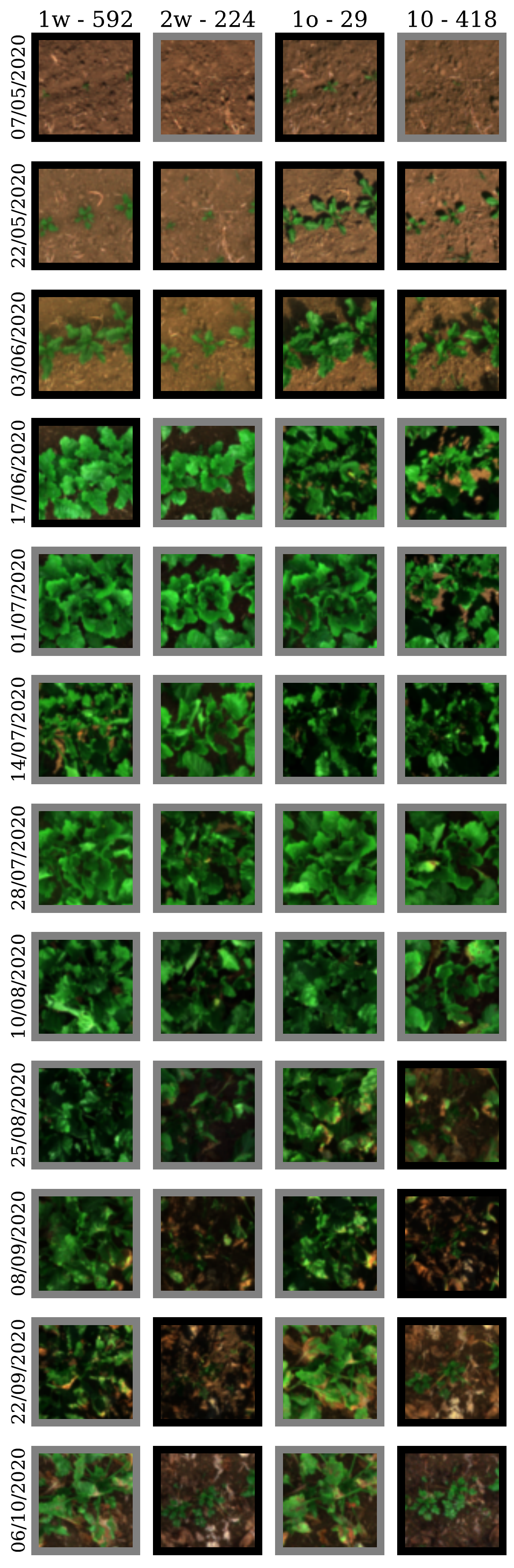}
    \caption{\textbf{Detected plant positions of sugar beet leaf spot dataset.} 4 randomly picked image series of plant RGB images detected by our method. The acquisition date increases downwards. Each second acquisition date is shown. Black frames annotate direct, gray frames indirect detections. The columns are ordered in blocks of 4 examples from inoculated, fungicide-treated, natural and reference fields, respectively.}
    \label{fig:cercospora_images_overview}
\end{figure}

Figure~\ref{fig:cercospora_images_overview} shows an image tile plot gathered with our workflow on the sugar beet dataset.

\end{document}